\documentclass{article}

\PassOptionsToPackage{numbers}{natbib}

\usepackage[preprint]{neurips_2024}

\usepackage[utf8]{inputenc} 
\usepackage[T1]{fontenc}    
\usepackage{hyperref}       
\usepackage{url}            
\usepackage{booktabs}       
\usepackage{amsmath}        
\usepackage{amsfonts}       
\usepackage{graphicx}
\usepackage{nicefrac}       
\usepackage{microtype}      
\usepackage{xcolor}         
\usepackage{subcaption}
\usepackage{multirow}

\newcommand{\totaltasks}{135}
\newcommand{\totalapps}{48}
\newcommand{\ourdataset}{Android-50}
\newcommand{\ourdomain}{autonomous UI agents}
\newcommand{\react}{ReAct}

\newcommand\secref[1]{Sec. \ref{#1}}
\newcommand\new[1]{{{#1}}}

\title{Latent State Estimation Helps UI Agents to Reason}

\author{%
  William E. Bishop \\
  Google Research
  \And
  Alice Li \\
  Google Research
  \And
  Christopher Rawles \\
  Google Research
  \And
  Oriana Riva \\
  Google Research
}

\begin{document}

\maketitle

\begin{abstract}
  A common problem for agents operating in real-world environments is that the response of an environment to their actions may be non-deterministic and observed through noise. This renders environmental state and progress towards completing a task latent. Despite recent impressive demonstrations of LLM's reasoning abilities on various benchmarks, whether LLMs can build estimates of latent state and leverage them for reasoning has not been explicitly studied. We investigate this problem in the real-world domain of \ourdomain. We establish that appropriately prompting LLMs in a zero-shot manner can be formally understood as forming point estimates of latent state in a textual space. In the context of autonomous UI agents we then show that LLMs used in this manner are more than $76\%$ accurate at inferring various aspects of latent state, such as performed (vs. commanded) actions and task progression. Using both public and internal benchmarks and three reasoning methods (zero-shot, CoT-SC \& \react), we show that LLM-powered agents that explicitly estimate and reason about latent state are able to successfully complete up to 1.6x more tasks than those that do not.
\end{abstract}

\section{Introduction}

\begin{figure}
\centering
\includegraphics[width=\columnwidth]{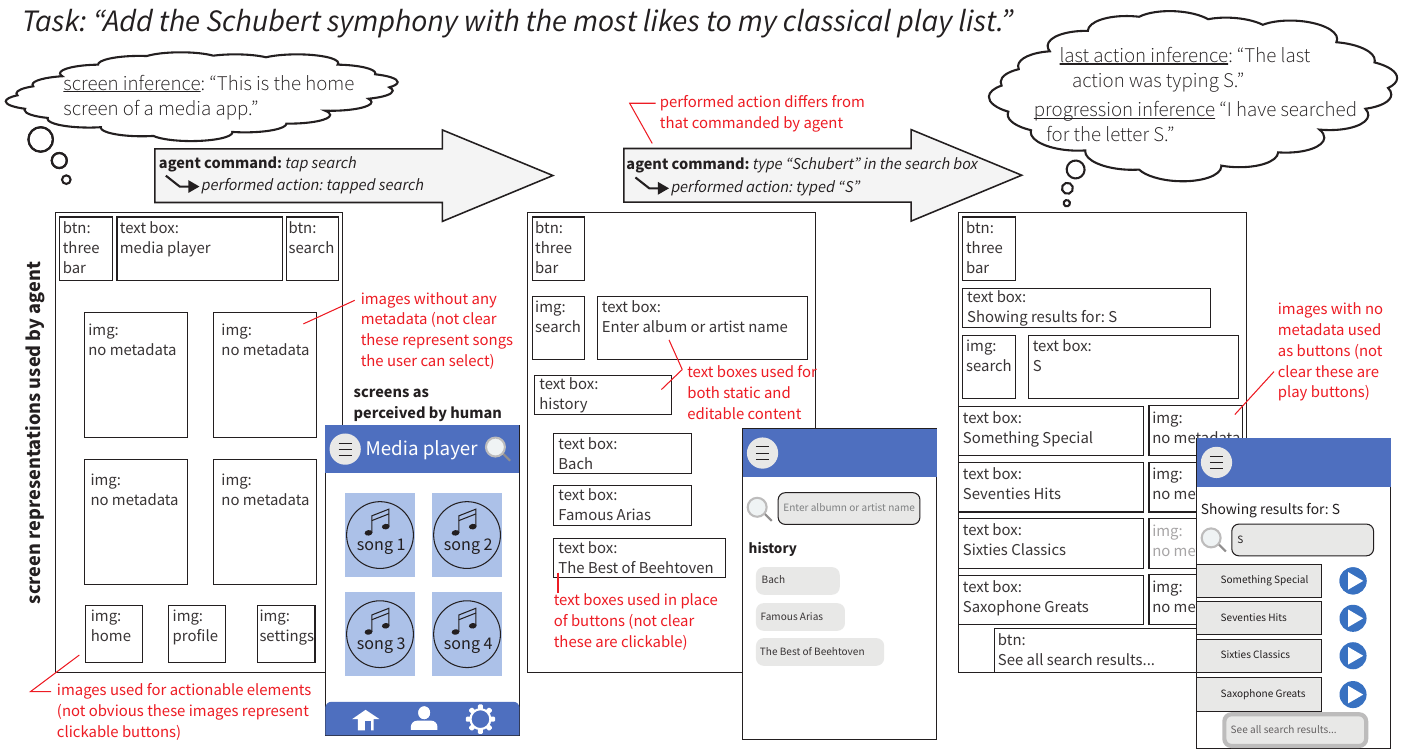}
\caption{A conceptual example of an agent performing a task while inferring latent state. The agent only perceives the screen through textual representations \new{(depicted as outlined diagrams for each screen) that may contain limited and noisy information about what is actually on the screen (shown in color in the lower right for each screen). Additionally, grounding errors (e.g., typing the wrong text into a box) can occur thus causing uncertainty about performed actions.} This means important aspects of UI and task state (e.g., high-level screen understanding, previously performed actions, progression from start) which can only be observed through these noisy representations are latent. We seek to explicitly estimate these and use these estimates to improve the selection of actions at each step.}
\label{fig:latent}
\end{figure}

While latent-state estimation plays a prominent role in many machine learning models for science and engineering~\cite{bollen2002latent, blei2014, cunningham2014dimensionality}, to the best of our knowledge the ability of LLMs to estimate latent state and use these estimates to improve decision making when performing tasks has not been studied. In this work, we investigate these problems in the emergent domain of LLM-based autonomous UI agents~\cite{li-acl20,yan2023gpt4v,zheng2023seeact}, in which agents freely interact with the user interface of an application or website (e.g., by clicking, typing and scrolling) to accomplish a variety of goals expressed in natural language. We choose this domain because it encompasses many real-world use cases and the number and types of tasks an agent is tested against can be easily scaled. Most importantly for our purposes, latent state also arises prominently (see Fig.~\ref{fig:latent}). This is because textual representations of screens are formed from noisy accessibility or DOM trees or the imperfect output of object detection models \cite{uied-icse20,screen-understanding-apple-chi21}. Additionally, performed actions may not match those commanded by an agent (due to grounding errors) or may have unexpected results. These factors render important aspects of UI and task state, such as what screen an agent is currently on, action outcomes, and progress towards a goal, latent.

In our approach, we seek to leverage the intuitive knowledge about the world encoded in LLMs to reason in a zero-shot manner from noisy observations about latent variables. \new{In the domain of UI automation}, this intuitive knowledge includes knowledge of user applications and their functionality, UI concepts, and task flows, and the latent variables we seek to infer include those useful for understanding where an agent is in an app and what it has accomplished towards a goal. Latent variable inference is challenging because there are often multiple values of the hidden variables that might explain the noisy and partial observations, and it is often only by reasoning across multiple observations and using prior modeled knowledge of the environment that reliable inferences can be made. We hypothesize that by prompting general pre-trained LLMs appropriately they can leverage the knowledge of the world encoded in their weights to assign high probabilities to completions that best explain observations, thereby performing latent variable estimation. Once some aspects of latent state are estimated, they can be provided in prompts to estimate other aspects of latent state and ultimately to select next actions to take, thus improving the performance of LLM-based agents.

Overall, we make the following contributions: \emph{(i)} we present a general method for forming point estimates of latent state in a zero-shot manner using LLMs, \emph{(ii)} we demonstrate that our methods can estimate five aspects of latent state for autonomous UI agents, matching or outperforming the performance of humans, and \emph{(iii)} we show that estimates of latent state can be naturally incorporated into existing reasoning techniques (zero-shot~\cite{wei2021finetuned}, CoT-SC~\cite{wang2022self}, \react~\cite{yao2022}) to consistently improve decision making for UI agents. Importantly, we establish these results on \emph{online} versions of three benchmarks: PixelHelp~\cite{li-acl20}, AndroidInTheWild~\cite{aitw2023}, and an internal benchmark, comprising in total \totaltasks~unique tasks from \totalapps~apps/websites. This is notable because, as observed by others~\cite{zheng2023seeact}, online testing is key to realistically assessing agent performance but is often skipped in favor of less realistic, but easier to implement, offline evaluation on pre-recorded data.

\section{Related Work}
\label{sec:related_work}

\paragraph{Latent state estimation}

Latent state arises in a variety of domains and modeling applications \cite[e.g.,][]{bollen2002latent, blei2014, cunningham2014dimensionality} where underlying variables of interest cannot be observed directly but instead only inferred from other variables.  Classic approaches to estimating latent state include principal component analysis \cite{pearson1901pca}, factor analysis \cite{spearman1904fa}, Kalman filtering \cite{kalman1960new}, and Hidden Markov models \cite{baum1966}, among others. Common to these and many other approaches is that latent state is modeled with continuous or discrete vectors and, before latent state can be inferred at test time, models must first be fit to application-specific training data.  Our work is different in that we seek to model latent state in a textual space by leveraging general, pre-trained LLMs without additional task-specific fine-tuning. With the exception of recent work that uses LLMs to infer politicians' latent positions~\cite{wu2023politics}, to the best of our knowledge our work is the first to formalize the general problem of using LLMs to estimate latent state and to show how this can be applied to improve autonomous UI agents. 

\paragraph{Reasoning with LLMs}
Our work fits into the broader research area of reasoning with LLMs. Increasingly sophisticated means of performing reasoning, decision making, and planning with LLMs have been proposed~\cite{wei2021finetuned,wang2022self,yao2022,zhou2022least, kojima-nips22, yao2023, wei2023chainofthought}. This line of work is predominantly concerned with the \emph{means} of reasoning. Our work is orthogonal to it as it is concerned with the \emph{content} of reasoning - that is showing that LLMs have the ability to reason about latent state. Having established this basic ability to reason over latent state, it is likely that absolute performance can be improved in the future by investigating different means of reasoning, using better base models, etc. 

\paragraph{\expandafter\MakeUppercase\ourdomain}

There is a rapidly growing body of work on UI agents~\cite{li-acl20,pmlr-v162-humphreys22a,pix2act,wang:chi2023,rci_kim2023language,yan2023gpt4v,zheng2023seeact,koh2024visualwebarena,webagent:iclr2024}. Our work is the first to explicitly recognize that important aspects of UI state and task progress are latent and to propose methods for LLM-based agents to address the challenges that arise from it.   

Additionally, our work differs from much of this prior work in two other ways.  First, our aim is \emph{not} to develop the best-performing agent, a goal which would require attention to many more aspects such as how UI screens are represented, how grounding is performed, what base models are used, etc. Instead, our goal is to show the relative improvements possible when agents explicitly incorporate reasoning about latent state in their planning, a general finding we believe can be used to improve other LLM-based agents. Second, most prior UI agent systems are tested only against pre-recorded datasets~\cite{bai2021uibert,wang:chi2023,yan2023gpt4v,aitw2023,motif,li-acl20,webagent:iclr2024}, with only a small number tested online~\cite{koh2024visualwebarena,he2024webvoyager,zheng2023seeact}. As explained in \S\ref{sec:methods}, testing online, where errors are allowed to accumulate, is critical for observing the benefits of latent state estimation for agent performance.

\paragraph{Reinforcement learning and robotics}

Formally, the problem of completing tasks by driving application UIs can be described as a partially-observable Markov decision process (POMDP) \cite{kaelbling1996reinforcement}, where the state of the UI and progress towards a goal are hidden variables.  While others have applied LLMs to related reinforcement learning problems~\cite{ahn2023, huang2022inner, liang2023code}, the novel aspect of our work is that we use LLMs to infer the hidden state of a system that can be modeled as a POMDP. 

\section{Estimating latent state with LLMs}
\label{sec:basic_theory}

We consider the problem of estimating the latent state of a system at time $t$, $s_t$, given a set of observations observed up until $t$, $\{o_i\}_{i=1}^t$.  A standard approach to forming a point estimate, $\hat{s}_t$, for latent state is to calculate
\vspace{-1ex}
\begin{align}
    \hat{s}_t = \text{argmax}_{s_t} p(s_t | \{o_i\}_{i=1}^t)
    \label{eq:general_problem}
\end{align}

\noindent Typically, $s_t$ and $o_t$ belong to Euclidean or discrete spaces, and models that are specifically fit using data collected for particular applications of interest are used to calculate the required probabilities. The key innovation of our work is to recognize that in many scenarios latent state and observations can be described in language. This opens up the possibility of using pre-trained LLMs to calculate the probabilities in \eqref{eq:general_problem}. 

The general approach can be formalized as follows. First, we assume that there is a set of $A$ different aspects of latent state \new{(e.g., high-level screen description, past actions, etc...)} we desire to estimate at each time $t$ that can be described in language.  We refer to the description of aspect $a$ at time $t$ as $s_t^a$, so that the full description of latent state at time $t$ is $s_t = \{s_t^a\}_{a=1}^A$.  At each time $t$, we use an LLM to estimate each aspect of latent state, and assume aspects are ordered in such a way that the estimate for one aspect (e.g., inferences of past actions, detected mistakes) are computed before and can inform the estimate for others (e.g., inference of progress towards a goal). \new{Specifically, at time $t$ we allow estimates of the $a^{th}$ aspect of latent state to depend on observations up until time $t$, $\{o_i\}_{i=1}^t$, estimates of latent state for all previous time steps, $\{s_{i}\}_{i=1}^{t-1}$, and estimates of aspects of latent state ordered before $a$ for the current time step, $\{s_t^i\}_{i=1}^{a-1}$.} For each aspect of latent state we then use a user-defined mapping (e.g., a heuristic), $f^a$, to form a prompt $z_t^a = f^a(\{o_i\}_{i=1}^t, \{s_i\}_{i=1}^{t-1}, \{s_t^i\}_{i=1}^{a-1})$.  Given this prompt, an LLM can be used to calculate the probabilities $p(s_t^a | z_t^a)$. Ideally, we would form a point estimate $\hat{s}_t^a = \text{argmax}_{s_t} p(s_t | z_t^a)$, but this is a computational prohibitive optimization, so in this work we approximate the mode with greedy decoding. Additional strategies such as sampling or beam search could be explored as well.

This general approach of using LLMs to estimate latent state has multiple benefits. First, it requires no application-specific training data. Instead, it leverages the ``intuitive knowledge'' LLMs have about the world to determine underlying state from noisy observations. Second, estimates of different aspects of latent state can be formed by simply using different prompts. Third, since they are expressed in language, latent state estimates are directly interpretable by humans. Finally, latent state estimated in this way can be fed to additional LLM calls (e.g., reasoning over actions to take). 

\section{Latent state estimation for UI automation}
\label{sec:latent_state_for_ui_automation}

\paragraph{Why does latent state arise in UI tasks?}

\begin{figure*}[t]
\centering
\begin{subfigure}[t]{0.48\columnwidth}
    \includegraphics[width=\columnwidth,trim=200 20 100 20,clip]{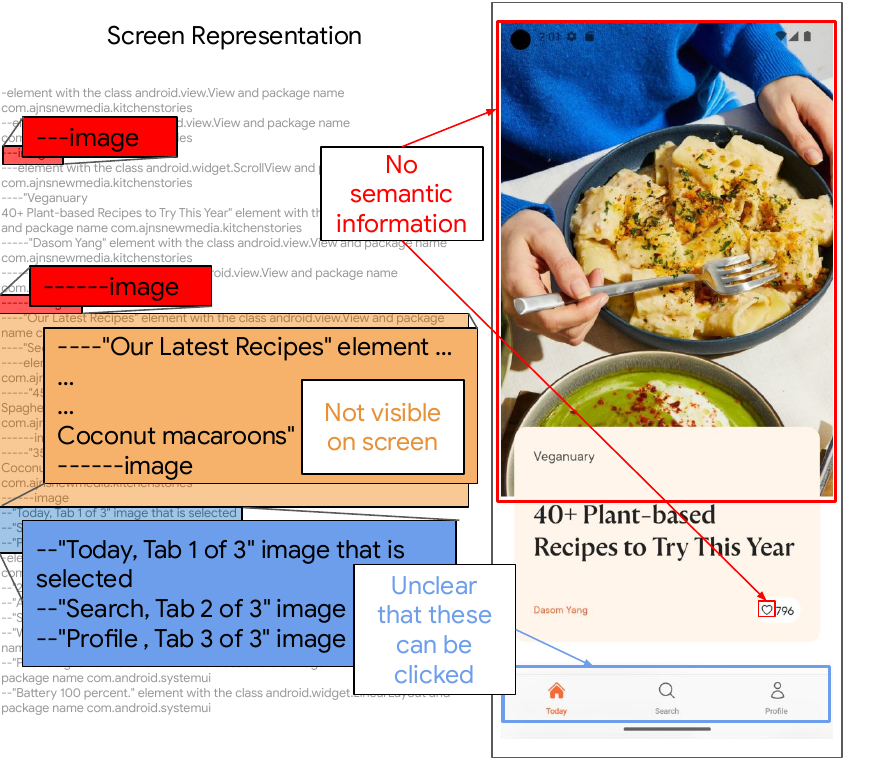}
    \caption{}
    \label{fig:screen-summary}
\end{subfigure}
\hfill
\begin{subfigure}[t]{0.48\columnwidth}
    \includegraphics[width=\columnwidth]{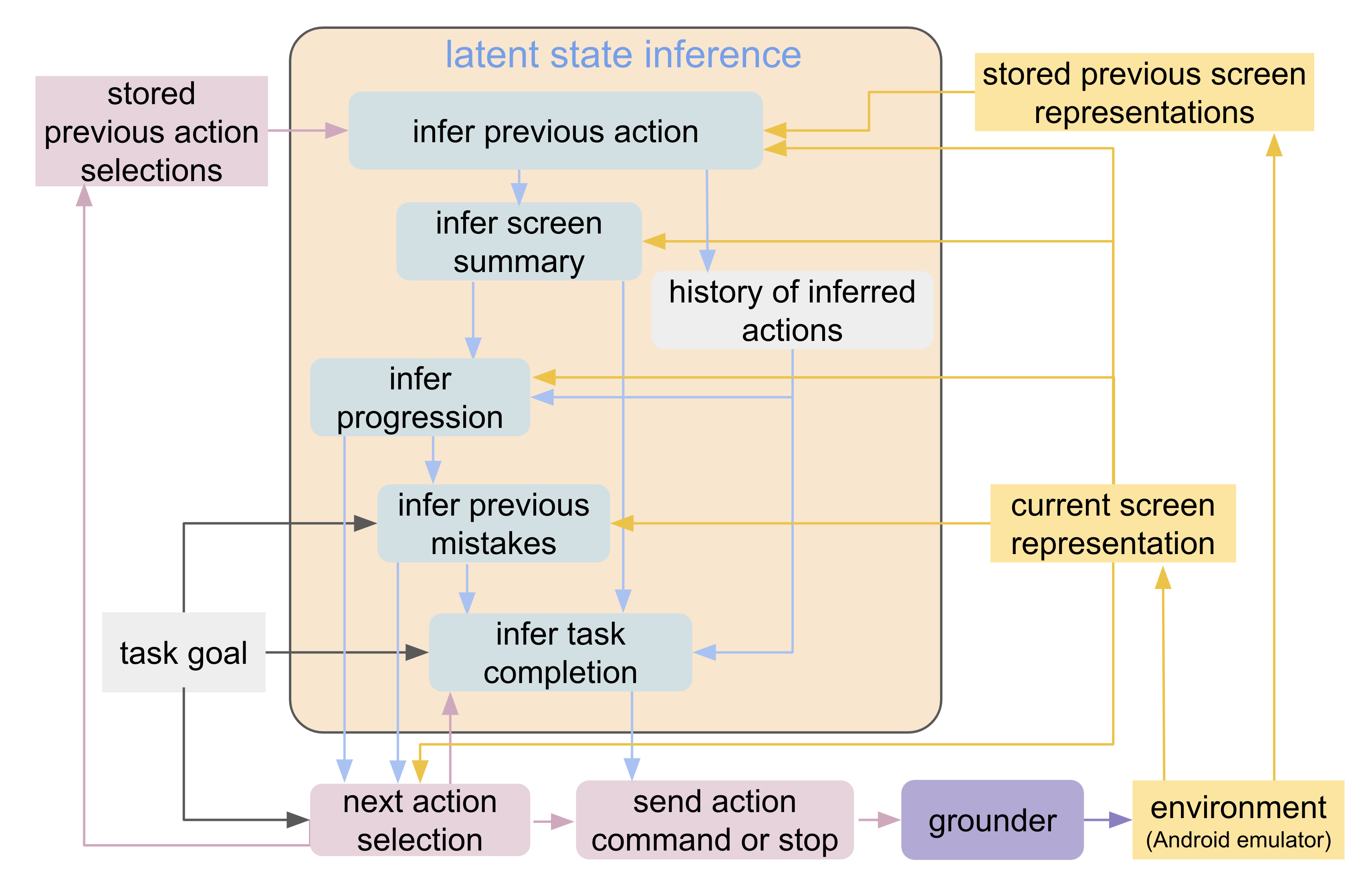}
     \caption{}
    \label{fig:estimates}
\end{subfigure}
\caption{(a) Examples of partial and noisy information in a representative screen description. The large image and the small heart icon at the bottom-right are represented as ``images'' without additional metadata. One third of the screen representation (in orange) refers to elements that are not visible on the screen. Finally, the three icons in the navigation bar are labeled simply as images, while they are actually clickable icons. (b) Process for inferring 5 aspects of latent state and selecting actions at each step of a task. Task completion is inferred after action selection, as we find providing contemplated next actions improves performance at detecting completion. The action is performed only if the the task is inferred as not done.}
\label{fig:screen-summary_and_latent_state}
\end{figure*}

Latent state arises in autonomous UI agents for at least two prominent reasons. First, observations, in the form of textual descriptions of screens and derived either from the output of object detection models or 
accessibility or DOM trees, are partial and noisy. In this work, we use descriptions derived from accessibility trees (see appendix \ref{app:forming-screen-repr} for details), but the issues we highlight are general. As shown in Fig.~\ref{fig:screen-summary_and_latent_state}a, in these descriptions elements may be missing key metadata necessary for understanding their content and function (partial information), background content invisible to the user may be present (noise) and the same type of elements, such as images, may be used as both actionable and static elements (so the element type is a noisy indicator of element function). Many other examples of how screen descriptions can be partial and noisy are provided in appendix \ref{app:ui-screens}. Second, actions selected by an agent may not be performed as expected, due to grounding errors. Grounding errors can occur in multiple forms such as failing to perform any action, performing an action (e.g,. clicking) on the wrong element, or performing the wrong action on the right element (e.g., typing the wrong thing in the right text field).  In addition to grounding errors, stochasticity in the environment can lead to unexpected action outcomes, such as the appearance of pop-up adds, an apparent screen freezing due to system delays or receiving unexpected dynamic content.

\new{In the context of UI automation,} we explore estimating $A=5$ aspects of latent state at each time step: (1) previous actions, (2) screen summaries, (3) progression, (4) previous mistakes, and (5) task completion. The above considerations mean all five of these aspects of state are latent. Previous actions are latent due to the unreliability of action grounding.  The noisy and partial nature of screen representations means that true screen content, and therefore quantities derived from them, such as screen summaries, are latent.  We define progression as a summary of what an agent has done from the start of a task (roughly analogous to integrating over actions at single time steps in classic latent dynamical system models~\cite{kalman1960new}). Progression, presence of previous mistakes, and task completion characterize in different ways the state of an agent's progress towards goal completion, and since they can only be inferred from noisy observations and unreliable action commands, are also latent. Fig.~\ref{fig:screen-summary_and_latent_state}b shows the order in which each aspect is estimated and how the estimates chain together.  \new{Relating this to the notation in \secref{sec:basic_theory}, at time $t$, each aspect corresponds to a different $s_t^a$ and is estimated with a separate prompt, $z_t^a$.} We provide the full LLM prompts in appendix \ref{app:latent_state_prompts}.%

\section{Methodology and benchmarks}
\label{sec:benchmarks}

\paragraph{Online e2e evaluation} We seek to evaluate our methods in the most realistic manner possible, and for this reason do not evaluate on pre-recorded datasets~\cite{li-acl20,aitw2023,yan2023gpt4v}. When testing on pre-recorded datasets, agents are shown the correct next screen, irrespective of their actions. This makes measuring realistic performance difficult, as inferring the state of a UI and progress towards a goal becomes harder as past mistakes in perception, decision making, and action execution accumulate, \new{which is preempted when testing on prerecorded data.}

Hence, we test our methods with three \emph{online} Android benchmarks. Agents are provided with a natural language goal. They then select their own actions and path through a UI on an emulated Android device using AndroidEnv \cite{android_env}, receiving observations in the form of textual descriptions of UI screens formed from the Android accessibility tree (appendix \ref{app:forming-screen-repr}) after each action. Agents perform a task until they determine it is complete or a termination criteria (max number of steps or 3 identical repeated actions) is reached. \new{The emulated devices used ran full versions of the Android operating system with the same application binaries that would be installed on real devices and agents received observations and performed actions in the same way they would on a physical device. For these reasons we are confident results reported here would translate to performance on physical devices.}

To test against a wide range of tasks and apps, we use three benchmarks. Two of the benchmarks, PixelHelp (PH) \cite{li-acl20} and Android In The Wild (AitW) \cite{aitw2023}, are formed by randomly selecting task goals from the corresponding datasets subject to minor constraints (e.g., ensuring tasks can still be feasibly performed in the present day). The third
benchmark, Android-50 (A-50), was collected by us using annotators (see appendix \ref{app:a50} for details). \new{A wide variety of real-world apps and tasks are present across this set of benchmarks so that a broad range of factors leading to partial and noisy screen representations and latent state in real-world use are well represented.} Table~\ref{tab:datasets} contains statistics for each benchmark, and a random subset of the tasks in each benchmark is provided in appendix \ref{sec:datasets}.  \new{AndroidEnv and both AitW and PH are all available under the Apache 2.0 license, and our use was in accordance with the license terms.}

\begin{table}
\centering
\caption{Statistics of benchmarks used in this work. Pooled numbers reflect that some apps are present in multiple benchmarks.}
\scalebox{1.0}{
\begin{tabular}{lcc}
\toprule 
 & \textbf{\# tasks} & \textbf{\# apps/websites} \\ 
\midrule 
PixelHelp (PH) & 35 & ~~4 \\ 
Android in the Wild (AitW) & 50 & 19 \\ 
\ourdataset~(A-50) & 50 & 33 \\ 
\midrule
\textbf{Total} & \textbf{135} & \textbf{48} \\ 
\bottomrule
\end{tabular}
}
\label{tab:datasets}
\end{table}

\paragraph{Agent architecture}
\label{sec:agent_architecture}
\label{sec:methods}

We used agents composed of two modules for (1) selecting the next action and (2) grounding selected actions in the current screen. Across experiments, we only varied (1) and held the grounder constant.  This meant that main factors that give rise to latent state, unfaithful screen representations and unreliable action execution, were constant across experiments, so we could directly compare the effects of varying the reasoning, including using latent state estimates or not, for selecting next actions. Grounding was performed with an LLM-based method adopting prior work~\cite{aitw2023,wang:chi2023} and supported clicking, typing, navigating backwards, and opening apps. See appendix \ref{app:grounder} for more details. 

\paragraph{Reasoning methods}

We implement three different methods for prompting LLMs to produce reasoning when selecting the next action of each task step:  \emph{zero-shot}~\cite{wei2021finetuned}, \emph{CoT-SC}~\cite{wang2022self} and \emph{\react}~\cite{yao2022} based reasoning. We implement two versions of each of these methods that either use (denoted by a +) or do not use (denoted by a -) latent state estimates. This allows us to perform pair-wise comparisons. We choose these three methods as they represent a well-known but diverse set of ways for eliciting reasoning from LLMs with various trade-offs such as the use of few-shot examples (CoT-SC \& \react) or not (zero-shot) as well as the amount of computation performed before selecting an action, with CoT-SC and \react~processing more tokens than zero-shot before predicting an action. 

Implementation details of all methods can be found in appendix \ref{app:action_selection}. In all cases, the textual description of the current screen and task goal was provided in the prompt. Estimates of progression and mistakes were additionally provided in the prompts of reasoning methods that used latent state, while agents without latent state only had access to the history of commanded actions (a common approach in UI agents~\cite{aitw2023, zheng2023seeact}). Stopping was based on estimates of completion for agents with latent state estimation, while agents without latent state estimation directly predicted stopping as an extra type of action. We note that given the length of our screen representations, we implement a modified version of \react~that retains only the last two previous observations and provides only a few action-thought pairs as few-shot examples at the beginning of the prompt (see \ref{sec:react_minus_prompt} and \ref{sec:react_plus_prompt} in the appendix for full prompts). Examples from three tasks, distinct from those in the benchmarks, were selected for inclusion in the prompts for CoT-SC and \react.  The same examples were used for each and were held fixed after they were selected, so as not to overfit to the benchmarks. We used a general pre-trained LLM, PaLM 2 text-unicorn \cite{palm2}, for all LLM calls. The temperature was set to 0, except when predicting actions with CoT-SC, in which case 8 samples were drawn with a temperature of .5.

\section{Results and analysis}
\label{sec:results}

\subsection{LLMs estimate latent state for UI agents}
\label{sec:latent_state_estimate_results}

\begin{table}[t]
\caption{Accuracy of the LLM at inferring various aspects of latent state across all benchmarks (left) and in comparison to humans (right).
}
\centering
\begin{tabular}{lcccc}
\toprule 
 & \multicolumn{4}{c}{evaluated on 40 tasks} \\
& \textbf{PH} & \textbf{AitW} & \textbf{A-50} & \textbf{total}  \\
\midrule 
  previous action & 90.0  & 77.6  & 95.2  & 89.4  \\
  screen summary & 90.0  & 93.4  & 91.2 & 91.5  \\
  progression & 94.3 & 72.4  & 91.8 & 87.4 \\
  previous mistakes & 80.0  & 68.4 & 79.6 & 76.8  \\
  task completion & 100.0  & 100.0  & 94.6 & 97.3  \\
\bottomrule
\end{tabular}
\quad
\begin{tabular}{cc}
\toprule 
\multicolumn{2}{c}{evaluated on 5 tasks}\\
 \textbf{LLM} & \textbf{human}  \\
\midrule 
   96.3 & 96.3  \\
   88.9 & 77.8  \\
   96.3 & 70.4 \\
   88.9 & 81.5  \\
  96.3 & 77.8 \\
\bottomrule
\end{tabular}
\label{tab:latent_state_estimation}
\end{table}

We now investigate the ability of our methods to estimate latent state for UI agents. We ran the zero-shot+ agent on each benchmark and we asked trained annotators to answer a series of ``yes'' or ``no'' questions to assess the accuracy of the estimate of each aspect of latent state listed in \S\ref{sec:latent_state_for_ui_automation} at each step. Example questions include ``Was the action that caused the transition between Step 1 and Step 2 inferred correctly?'' or ``Does the mistake assessment correctly capture the mistakes (if any) that have been made up to Step 1 and are not yet corrected?''  The full set of questions can be found in appendix \ref{app:latent_state-questions}. \new{Annotators used throughout this study signed a data usage agreement, were fairly compensated and provided no personal data as part of this study.} This is a costly analysis, so we only performed this on a random subset of 10 tasks from PixelHelp, 10 tasks from AitW, and 20 from \ourdataset.  We sampled more from \ourdataset~because of its greater diversity of apps and websites. Cost is also the reason we only analyzed latent state estimates produced by the zero-shot+ agent. We believe this is justified because the chain of LLM calls for estimating latent state is the same across all tested methods for selecting next actions. 

Table~\ref{tab:latent_state_estimation} summarizes our findings. We observe that the basic prompting strategy outlined in \S\ref{sec:latent_state_for_ui_automation} estimated all five aspects of latent state with remarkably high accuracy, i.e., 76.8\%--97.3\%.  Three aspects of latent state: previously performed actions, mistakes, and completion, could potentially be estimated with relatively high accuracy by na\"ive \new{baselines} that always predict the defaults, i.e., the performed action matches the commanded one, no mistakes are made, and the task is not done for all steps. We determined that in all cases these \new{na\"ive baselines} would achieve lower performance (previous action accuracy of $85\%$, a mistake assessment accuracy of $74.0\%$, and a task completion accuracy of $92.8\%$) than that achieved by the LLM. We also inspected the ``hard'' cases for each of these aspects, which occurred when performed actions did not match commanded actions, when mistakes occurred, and for the steps when the task should be marked as completed. In these cases, we found that the LLM correctly predicted the previous action $61.3\%$ of the time, correctly detected and described mistakes $25.9\%$ of the time, and correctly predicted completion $81.0\%$ of the time. For further details on how these values were computed see appendix \ref{app:naive-method}.

\begin{table*}[t]
\centering
\caption{Performance (as percentages) on all three benchmarks for UI agents which incorporated (denoted by +) or did not incorporated (denoted by -) estimates of latent state in selecting actions at each step. }
\scalebox{0.7}{
\begin{tabular}{lcccccccccccccccc}
\toprule
 &  \multicolumn{4}{c}{\textbf{task success} $\uparrow$} & \multicolumn{4}{c}{\textbf{partial completion} $\uparrow$}  & \multicolumn{4}{c}{\textbf{task success with strict stop} $\uparrow$} & \multicolumn{4}{c}{\textbf{premature stop} $\downarrow$} \\
 & \textbf{PH} & \textbf{AitW} & \textbf{A-50} & \textbf{tot} & \textbf{PH} & \textbf{AitW} & \textbf{A-50} & \textbf{tot} & \textbf{PH} & \textbf{AitW} & \textbf{A-50} & \textbf{tot} & \textbf{PH} & \textbf{AitW} & \textbf{A-50} & \textbf{tot}\\
 \midrule
 zero-shot + & 62.9 & 46.0 & 34.0 & 45.9 & 73.7 & 61.2 & 57.0 & 62.1 & 62.9 & 40.0 & 28.0 & 41.5 & 11.4 & 12.0 & 24.0 & 16.3\\ 
 zero-shot - & 45.7 & 24.0 & 20.0 & 28.1 & 66.3 & 45.2 & 48.2 & 51.0 & 45.7 & 18.0 & 20.0 & 25.9 & 54.3 & 68.0 & 80.0 & 68.9 \\ \midrule
 CoT-SC + & 62.9 & 42.0 & 28.0 & 42.2 & 82.0 & 53.0 & 48.7 & 57.5 & 48.6 & 32.0 & 22.0 & 32.6 & 22.9 & 20.0 & 30.0 & 24.4 \\ 
 CoT-SC - & 42.9 & 26.0 & 20.0 & 28.1 & 68.0 & 42.8 & 43.5 & 48.6 & 42.9 & 24.0 & 20.0 & 27.4 & 40.0 & 52.0 & 66.0 & 54.1 \\ \midrule
 \react + & 68.6 & 44.0 & 26.0 & 43.7 & 83.0 & 58.4 & 54.4 & 62.1 & 65.7 & 32.0 & 24.0 & 37.8 & 8.6 & 12.0 & 20.0 & 14.1\\ 
 \react~- & 60.0 & 28.0 & 26.0 & 35.6 & 76.0 & 46.4 & 45.6 & 52.5 & 57.1 & 22.0 & 26.0 & 32.6 & 17.1 & 6.0 & 12.0 & 11.1\\ 
 \bottomrule
\end{tabular}
}
\label{tab:e2e}
\label{tab:stopping}
\end{table*}

We also asked five human volunteers (not authors of this paper), who are experts in Android programming and fluent in English, to produce estimates of latent state as the zero-shot+ agent performed five randomly selected tasks ($27$ steps in total).  We ensured the volunteers only had access to the same information the LLMs did by providing them with the same prompts.  To minimize the possibility that differences in performance were due to the difficulty humans might have in parsing long blocks of text, we colored coded the screen representations in the prompts to aid in visual understanding, but beyond this provided no additional information (e.g., screenshots) to the humans. Just as with the LLM-produced estimates, previous human estimates of aspects of latent state (e.g., past inferred actions) were fed into the prompts for other aspects of latent state in the same and later steps. Finally, to minimize the possibility of humans using memory from previous steps the LLMs did not have access to, we ensured no volunteer ever estimated an aspect of latent state for screens less than 4 steps apart. The volunteers were informed of the intended use of the data and no personal data was collected.

As shown in Table~\ref{tab:latent_state_estimation}, humans achieved high, but imperfect performance, underscoring the difficulty of estimating latent state in this setting.  \new{Using paired permutation tests, we found LLMs significantly outperformed humans when estimating progression ($p = .02$), moderately outperformed when estimating task completion ($p = .06$), and that there was no significant difference between humans and LLMs for the other aspects of latent state.  The ability of LLMs to match or outperform humans is evidence of their strong performance on this task.}

\subsection{Reasoning about latent state improves UI agent performance}
\label{sec:e2e_results}

We next evaluate how the performance of UI agents can improve when estimates of latent state are used when selecting actions at each step. Due to the multiple possible paths agents may take to accomplish a goal on our emulators, automated evaluation is difficult.  Therefore, to measure performance we ask annotators to answer a series of ``yes'' or ``no'' questions to assess (1) whether an agent accomplished a task (irrespective of whether it stopped correctly or not), (2) whether it was able to stop, and (3) how many of the sub-tasks involved in completing a full task the agent performed correctly. The last is a measure of ``partial completion'' which we obtain through questions manually curated for each task (e.g., "Did the agent open the correct app?", "Did the agent perform a search?", etc.). The number of questions assessing partial completion varied from 1 to 7, depending on the task complexity. A sample of the questions asked can be found in appendix \ref{app:rater-questions}.

As shown in Table~\ref{tab:e2e}, providing estimates of latent state when reasoning about next actions to take, substantially improves performance of UI agents.  This was uniformly true when measuring task success and partial completion at both the pooled and individual benchmark level. \new{The percentage of tasks that were successfully completed (see ``task success'' columns in the table), measured in total across all benchmarks, increased by factors of 1.6 (from 28.1\% to 45.9\%, p = .0002), 1.5 (from 28.1\% to 42.2\% p = .0041) and 1.2 (from 35.6\% to 43.7\%, p = .078) respectively, for zero-shot, CoT-SC, and \react~based methods with latent state estimates incorporated.} \new{Here and throughout this section, p-values are for the difference in performance for each reasoning method, and were calculated with two-sided paired-permutation tests.} In addition, whether tasks were successfully completed or not, overall progress achieved toward task goals, as measured by partial completion, similarly improved when latent state estimates were used \new{(improvements in partial completion for all reasoning methods were all significant with p < .001)}. This is strong evidence that LLM-powered UI agents, irrespective of \emph{how} they perform reasoning, benefit from incorporating the explicit estimation of the \emph{content} of latent state into their reasoning process.

The benefits of incorporating latent state estimates can be seen in other aspects of agent performance as well, such as \emph{stopping}.  An ideal agent should stop immediately after a task is complete without taking extra steps, while at the same time not stopping prematurely.  Using the annotators' responses, we can measure both of these aspects of agent performance (see third and forth sets of columns in Table~\ref{tab:stopping}). We quantify the percentage of tasks in which the agent not only accomplished the goal but stopped without taking extra steps. This is a stricter measure than simple task success, which tolerates extra steps (e.g., navigating back to a home screen, clicking on a specific result after performing a search, etc.), and it is likely to be associated with user satisfaction. We find that similar to the more lenient measure, task success measured with this stricter stopping criteria improves by factors of \new{1.6 (from 25.9\% to 41.5\%, p = .0009), 1.2 (from 27.4\% to 32.6\%, p=.31) and 1.2 (from 32.6\% to 37.8\%, p=.31) respectively, for agents using zero-shot, CoT-SC and \react~based methods with latent state estimates incorporated.} \new{Additionally, the zero-shot and CoT-SC agents without latent state estimation stopped prematurely 4.2 times (68.9\% to 16.3\% of tasks, p < .0001) and 2.2 times (54.1\% to 24.4\% of tasks, p < .0001) more often then versions with latent state estimation, while the performance of the \react~agent with latent state estimation decreased only slightly in this regard, stopping prematurely 1.3 times (14.1\% to 11.1\%, p = .59) more than the agent without latent state.} \new{Upon investigation we discovered this did not reflect superior stopping performance of the \react~- agent, but that the \react~- agent often failed to stop at all (and therefore rarely prematurely stopped). In particular, we found the \react~- agent only stopped in 37.8\% of episodes while \react~+ stopped in 57.5\% of episodes before reaching the maximum number of steps we allowed.  In addition, by analyzing only episodes where agents stopped before reaching the max number of steps, we found that the \react~- agent had worse premature stopping performance (when it
stopped, 33.3\% of the time it was premature) than \react~+ (when it stopped, 29.9\% of the time it was premature).} 

\subsection{Error analysis}
The high accuracy achieved at estimating latent state (\S\ref{sec:latent_state_estimate_results}), did not translate into similarly high end-to-end performance.  One explanation is that results for latent state estimation ( Table~\ref{tab:latent_state_estimation}) are quantified across single steps of tasks, and to perform a task successfully agents must perform multiple steps. Each step introduces the possibility that one mistake in latent state estimation (e.g., determining a task is prematurely complete) could adversely affect task performance. Additional factors, such as grounding performance and other aspects of reasoning not studied here (e.g., more advanced capabilities for exploring apps when one execution path fails), also affect performance.

\begin{figure}[t]
\centering
\includegraphics[width=.5\columnwidth,trim=20 75 75 45,clip]{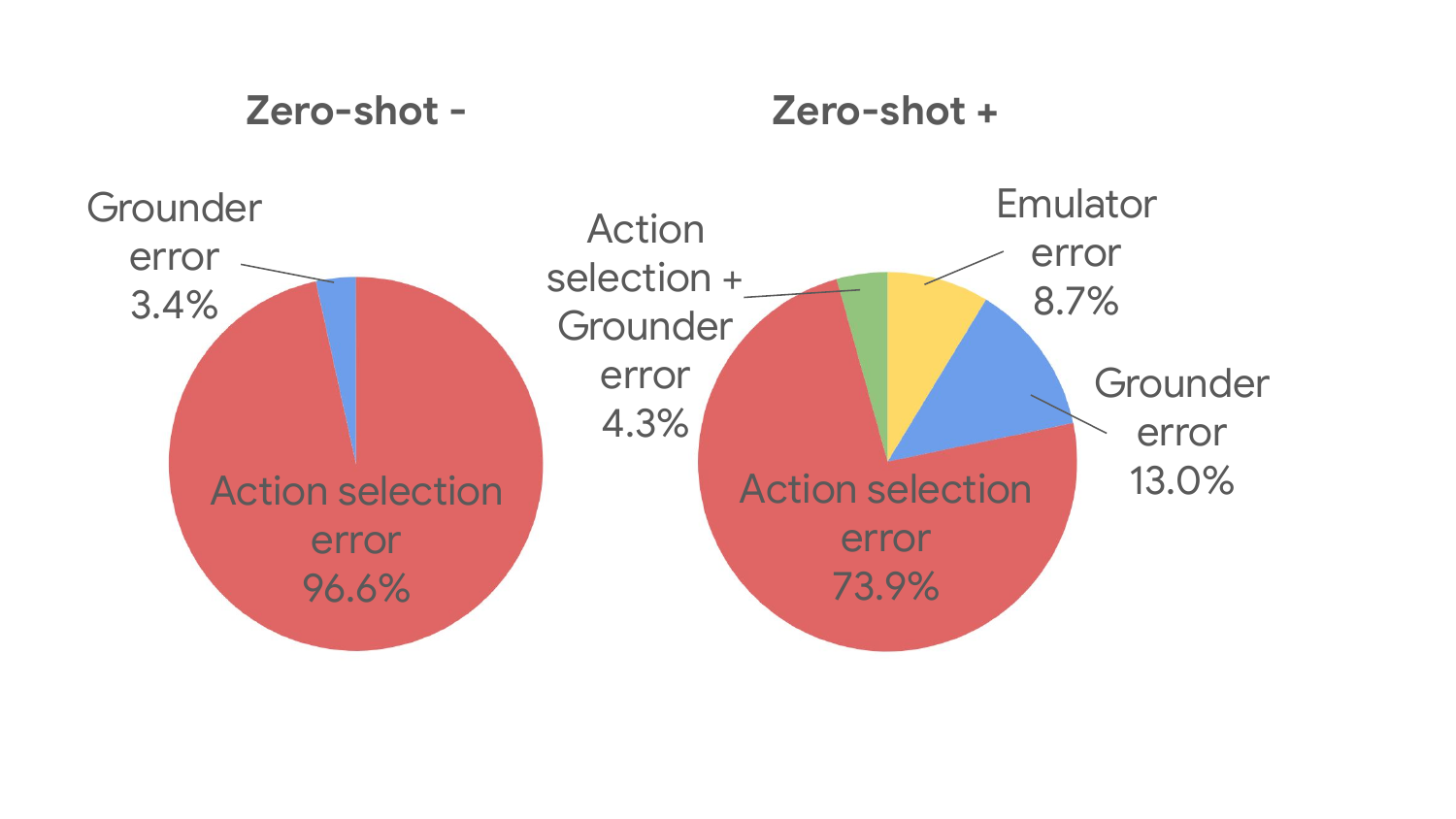}
\vspace{-1ex}
\caption{Distribution of failure causes for agents with zero-shot action selection without and with latent state.}
\label{fig:error_analysis}
\end{figure}

Nonetheless, if latent state is useful for planning next steps, then the cause of failure should shift away from action selection to other causes, such as grounding mistakes. We assessed this through an error analysis of tasks the agents with zero-shot action selection failed on. We manually inspected both predicted actions and grounder behavior and categorized the root cause of a failed task as being due either to action selection (e.g., selecting wrong actions for a task) or grounding (selected actions performed incorrectly). While there were often multiple mistakes per task, it was often possible to judge the primary reason. For completeness, we also noted rare cases when the cause could have been due to both or was due to the the emulator (e.g., failing to load an app). Due to the effort involved, we only examined failed tasks that were part of the same 40 examined in \S\ref{sec:latent_state_estimate_results} (31 failed tasks for zero-shot- and 26 for zero-shot+). As shown in Fig.~\ref{fig:error_analysis}, the percentage of tasks that failed due to action selection decreased substantially when latent state estimates were used in action selection, indicating that as latent state improves reasoning for next actions, other weaknesses, such as grounding, start to play a bigger role in bottlenecking task success. 

Finally, while our goal was to compare the same methods with and without latent state, we briefly comment on the relative performance of the three reasoning methods. The tasks we test against are very diverse, and it has been observed few-shot examples may not always generalize well  and performance can be quite sensitive to their ordering and wording \cite{min2022rethinking, balashankar2023, zhao2021calibrate, lu2021fantastically}. While we could have tuned our prompts to improve performance, we avoided doing this so as not to unfairly fit to our test data. It is also interesting to note that while including latent state improved performance of all methods, best performance was obtained with the zero-shot method.  This suggests the possibility that once key aspects of latent state are estimated and can be provided in a prompt, simple reasoning without few-shot examples may be sufficient for predicting actions. This is intriguing as it suggests a way of breaking through the need to provide many examples in a prompt, which may be impossible or very costly when a large variety of tasks must be performed, to achieve robust and general UI agents.

\section{Conclusions}
\label{sec:conclusions}

We have proposed a general approach for using LLMs to estimate latent state, and demonstrated its effectiveness in estimating five aspects of latent state for UI agents. We showed how these estimates can be incorporated into reasoning about next action prediction, leading to an increases of up to 1.6x in LLM-agents performing tasks end-to-end in emulated Android environments.  We believe the findings that LLMs can estimate latent state for UI agents and that latent state helps with their decision making likely generalize to other POMDP-style domains.

\paragraph{Limitations}

Our UI agent implementation is based on only one platform (Android) but we expect results to generalize to others, especially web.  While we also tested only with one method of forming screen representations, we expect screen understanding, including that by vision-language models (VLM), to be imperfect going into the foreseeable future, so that latent UI and task states, as considered here, will continue to arise. For this reason, we believe the general approach of incorporating the estimation of latent state into agents to be broadly applicable going forward. We only ran with a single LLM, but we expect results to generalize to others and end-to-end performance to increase with newer models.  The cost of using human annotators for evaluation kept the total number of tasks we could evaluate on modest. Nonetheless, the number of tasks we report on is inline with other recent work that evaluates online. For cost reasons, we also only considered a fixed set of reasoning methods for selecting next actions. We could choose others (e.g., tree-of-thoughts~\cite{yao2023}) but given that we see uniform improvements across all methods we tested, we expect our general finding (that performance improves by incorporating estimates of latent state) to hold. 

\paragraph{Broader impacts}

UI agents can benefit visually-impaired users, by providing them with access to a much wider range of applications and functionality. Another potential use case is general task automation, which has societal, security and privacy implications. An agent may leak private information or carry out a task in an unacceptable way or produce unwanted side effects. Malicious actors could also use UI agents for undesired purposes such as overriding anti-fraud mechanisms or manipulating applications to achieve undesirable goals. For these reasons, deployment of this technology going forward will have to be carefully considered and combined with research in other areas on LLM safety to balance potential societal benefits with risks.

\newpage
\bibliographystyle{abbrv}
\bibliography{}

\begin{thebibliography}{10}

\bibitem{ahn2023}
M.~Ahn, A.~Brohan, N.~Brown, Y.~Chebotar, O.~Cortes, B.~David, C.~Finn, C.~Fu,
  K.~Gopalakrishnan, K.~Hausman, et~al.
\newblock Do as i can, not as i say: grounding language in robotic affordances.
\newblock In {\em Conference on Robot Learning}, pages 287--318. PMLR, 2023.

\bibitem{bai2021uibert}
C.~Bai, X.~Zang, Y.~Xu, S.~Sunkara, A.~Rastogi, J.~Chen, and B.~A. y~Arcas.
\newblock {UIBert}: Learning generic multimodal representations for {UI}
  understanding.
\newblock In Z.~Zhou, editor, {\em Proc. of the 30th International Joint
  Conference on Artificial Intelligence, {IJCAI} 2021}, pages 1705--1712.
  ijcai.org, 2021.

\bibitem{balashankar2023}
A.~Balashankar, X.~Ma, A.~Sinha, A.~Beirami, Y.~Qin, J.~Chen, and A.~Beutel.
\newblock Improving few-shot generalization of safety classifiers via data
  augmented parameter-efficient fine-tuning.
\newblock {\em arXiv preprint arXiv:2310.16959}, 2023.

\bibitem{baum1966}
L.~E. Baum and T.~Petrie.
\newblock Statistical inference for probabilistic functions of finite state
  markov chains.
\newblock {\em The annals of mathematical statistics}, 37(6):1554--1563, 1966.

\bibitem{blei2014}
D.~M. Blei.
\newblock Build, compute, critique, repeat: Data analysis with latent variable
  models.
\newblock {\em Annual Review of Statistics and Its Application}, 1:203--232,
  2014.

\bibitem{bollen2002latent}
K.~A. Bollen.
\newblock Latent variables in psychology and the social sciences.
\newblock {\em Annual review of psychology}, 53(1):605--634, 2002.

\bibitem{motif}
A.~Burns, D.~Arsan, S.~Agrawal, R.~Kumar, K.~Saenko, and B.~A. Plummer.
\newblock Mobile app tasks with iterative feedback (motif): Addressing task
  feasibility in interactive visual environments.
\newblock {\em CoRR}, abs/2104.08560, 2021.

\bibitem{chen20:gui-accessibility}
J.~Chen, C.~Chen, Z.~Xing, X.~Xu, L.~Zhu, G.~Li, and J.~Wang.
\newblock {Unblind Your Apps: Predicting Natural-Language Labels for Mobile GUI
  Components by Deep Learning}.
\newblock In {\em Proc. of the ACM/IEEE 42nd International Conference on
  Software Engineering}, ICSE '20, pages 322--334, 2020.

\bibitem{uied-icse20}
J.~Chen, M.~Xie, Z.~Xing, C.~Chen, X.~Xu, L.~Zhu, and G.~Li.
\newblock Object detection for graphical user interface: Old fashioned or deep
  learning or a combination?
\newblock In {\em Proc. of the 28th ACM Joint Meeting on European Software
  Engineering Conference and Symposium on the Foundations of Software
  Engineering}, ESEC/FSE 2020, pages 1202--1214, 2020.

\bibitem{cunningham2014dimensionality}
J.~P. Cunningham and B.~M. Yu.
\newblock Dimensionality reduction for large-scale neural recordings.
\newblock {\em Nature neuroscience}, 17(11):1500--1509, 2014.

\bibitem{palm2}
R.~A. Google~and, A.~M. Dai, O.~Firat, M.~Johnson, D.~Lepikhin, A.~Passos,
  S.~Shakeri, E.~Taropa, P.~Bailey, Z.~Chen, E.~Chu, J.~H. Clark, L.~E. Shafey,
  Y.~Huang, K.~Meier-Hellstern, G.~Mishra, E.~Moreira, M.~Omernick,
  K.~Robinson, S.~Ruder, Y.~Tay, K.~Xiao, Y.~Xu, Y.~Zhang, G.~H. Abrego,
  J.~Ahn, J.~Austin, P.~Barham, J.~Botha, J.~Bradbury, S.~Brahma, K.~Brooks,
  M.~Catasta, Y.~Cheng, C.~Cherry, C.~A. Choquette-Choo, A.~Chowdhery,
  C.~Crepy, S.~Dave, M.~Dehghani, S.~Dev, J.~Devlin, M.~Díaz, N.~Du, E.~Dyer,
  V.~Feinberg, F.~Feng, V.~Fienber, M.~Freitag, X.~Garcia, S.~Gehrmann,
  L.~Gonzalez, G.~Gur-Ari, S.~Hand, H.~Hashemi, L.~Hou, J.~Howland, A.~Hu,
  J.~Hui, J.~Hurwitz, M.~Isard, A.~Ittycheriah, M.~Jagielski, W.~Jia,
  K.~Kenealy, M.~Krikun, S.~Kudugunta, C.~Lan, K.~Lee, B.~Lee, E.~Li, M.~Li,
  W.~Li, Y.~Li, J.~Li, H.~Lim, H.~Lin, Z.~Liu, F.~Liu, M.~Maggioni,
  A.~Mahendru, J.~Maynez, V.~Misra, M.~Moussalem, Z.~Nado, J.~Nham, E.~Ni,
  A.~Nystrom, A.~Parrish, M.~Pellat, M.~Polacek, A.~Polozov, R.~Pope, S.~Qiao,
  E.~Reif, B.~Richter, P.~Riley, A.~C. Ros, A.~Roy, B.~Saeta, R.~Samuel,
  R.~Shelby, A.~Slone, D.~Smilkov, D.~R. So, D.~Sohn, S.~Tokumine, D.~Valter,
  V.~Vasudevan, K.~Vodrahalli, X.~Wang, P.~Wang, Z.~Wang, T.~Wang, J.~Wieting,
  Y.~Wu, K.~Xu, Y.~Xu, L.~Xue, P.~Yin, J.~Yu, Q.~Zhang, S.~Zheng, C.~Zheng,
  W.~Zhou, D.~Zhou, S.~Petrov, and Y.~Wu.
\newblock Palm 2 technical report, 2023.

\bibitem{webagent:iclr2024}
I.~Gur, H.~Furuta, A.~Huang, M.~Safdari, Y.~Matsuo, D.~Eck, and A.~Faust.
\newblock A real-world webagent with planning, long context understanding, and
  program synthesis.
\newblock {\em arXiv preprint arXiv:2307.12856}, 2023.

\bibitem{he2024webvoyager}
H.~He, W.~Yao, K.~Ma, W.~Yu, Y.~Dai, H.~Zhang, Z.~Lan, and D.~Yu.
\newblock Webvoyager: Building an end-to-end web agent with large multimodal
  models.
\newblock {\em arXiv preprint arXiv:2401.13919}, 2024.

\bibitem{huang2022inner}
W.~Huang, F.~Xia, T.~Xiao, H.~Chan, J.~Liang, P.~Florence, A.~Zeng, J.~Tompson,
  I.~Mordatch, Y.~Chebotar, et~al.
\newblock Inner monologue: Embodied reasoning through planning with language
  models.
\newblock {\em arXiv preprint arXiv:2207.05608}, 2022.

\bibitem{pmlr-v162-humphreys22a}
P.~C. Humphreys, D.~Raposo, T.~Pohlen, G.~Thornton, R.~Chhaparia, A.~Muldal,
  J.~Abramson, P.~Georgiev, A.~Santoro, and T.~Lillicrap.
\newblock A data-driven approach for learning to control computers.
\newblock In K.~Chaudhuri, S.~Jegelka, L.~Song, C.~Szepesvari, G.~Niu, and
  S.~Sabato, editors, {\em Proceedings of the 39th International Conference on
  Machine Learning}, volume 162 of {\em Proceedings of Machine Learning
  Research}, pages 9466--9482. PMLR, 17--23 Jul 2022.

\bibitem{kaelbling1996reinforcement}
L.~P. Kaelbling, M.~L. Littman, and A.~W. Moore.
\newblock Reinforcement learning: A survey.
\newblock {\em Journal of artificial intelligence research}, 4:237--285, 1996.

\bibitem{kalman1960new}
R.~E. Kalman et~al.
\newblock A new approach to linear filtering and prediction problems [j].
\newblock {\em Journal of basic Engineering}, 82(1):35--45, 1960.

\bibitem{rci_kim2023language}
G.~Kim, P.~Baldi, and S.~M. McAleer.
\newblock Language models can solve computer tasks.
\newblock In {\em Thirty-seventh Conference on Neural Information Processing
  Systems}, 2023.

\bibitem{koh2024visualwebarena}
J.~Y. Koh, R.~Lo, L.~Jang, V.~Duvvur, M.~C. Lim, P.-Y. Huang, G.~Neubig,
  S.~Zhou, R.~Salakhutdinov, and D.~Fried.
\newblock Visualwebarena: Evaluating multimodal agents on realistic visual web
  tasks.
\newblock {\em arXiv preprint arXiv:2401.13649}, 2024.

\bibitem{kojima-nips22}
T.~Kojima, S.~S. Gu, M.~Reid, Y.~Matsuo, and Y.~Iwasawa.
\newblock Large language models are zero-shot reasoners.
\newblock In S.~Koyejo, S.~Mohamed, A.~Agarwal, D.~Belgrave, K.~Cho, and A.~Oh,
  editors, {\em Advances in Neural Information Processing Systems}, volume~35,
  pages 22199--22213. Curran Associates, Inc., 2022.

\bibitem{li2023spotlight}
G.~Li and Y.~Li.
\newblock Spotlight: Mobile {UI} understanding using vision-language models
  with a focus.
\newblock In {\em Proc. of the 11th International Conference on Learning
  Representations (ICLR)}, 2023.

\bibitem{li-acl20}
Y.~Li, J.~He, X.~Zhou, Y.~Zhang, and J.~Baldridge.
\newblock Mapping natural language instructions to mobile {UI} action
  sequences.
\newblock In {\em Proc. of the 58th Annual Meeting of the Association for
  Computational Linguistics, {ACL} 2020, Online, July 5-10, 2020}, pages
  8198--8210. Association for Computational Linguistics, 2020.

\bibitem{liang2023code}
J.~Liang, W.~Huang, F.~Xia, P.~Xu, K.~Hausman, B.~Ichter, P.~Florence, and
  A.~Zeng.
\newblock Code as policies: Language model programs for embodied control.
\newblock In {\em 2023 IEEE International Conference on Robotics and Automation
  (ICRA)}, pages 9493--9500. IEEE, 2023.

\bibitem{lu2021fantastically}
Y.~Lu, M.~Bartolo, A.~Moore, S.~Riedel, and P.~Stenetorp.
\newblock Fantastically ordered prompts and where to find them: Overcoming
  few-shot prompt order sensitivity.
\newblock {\em arXiv preprint arXiv:2104.08786}, 2021.

\bibitem{min2022rethinking}
S.~Min, X.~Lyu, A.~Holtzman, M.~Artetxe, M.~Lewis, H.~Hajishirzi, and
  L.~Zettlemoyer.
\newblock Rethinking the role of demonstrations: What makes in-context learning
  work?
\newblock {\em arXiv preprint arXiv:2202.12837}, 2022.

\bibitem{pearson1901pca}
K.~Pearson.
\newblock Liii. on lines and planes of closest fit to systems of points in
  space.
\newblock {\em The London, Edinburgh, and Dublin philosophical magazine and
  journal of science}, 2(11):559--572, 1901.

\bibitem{aitw2023}
C.~Rawles, A.~Li, D.~Rodriguez, O.~Riva, and T.~Lillicrap.
\newblock Android in the wild: A large-scale dataset for android device
  control.
\newblock In {\em NeurIPS 2023 Datasets and Benchmarks Track}, 2023.

\bibitem{pix2act}
P.~Shaw, M.~Joshi, J.~Cohan, J.~Berant, P.~Pasupat, H.~Hu, U.~Khandelwal,
  K.~Lee, and K.~Toutanova.
\newblock From pixels to {UI} actions: Learning to follow instructions via
  graphical user interfaces.
\newblock In {\em Thirty-seventh Conference on Neural Information Processing
  Systems}, 2023.

\bibitem{spearman1904fa}
C.~Spearman.
\newblock General intelligence, objectively determined and measured.
\newblock {\em American Journal of Psychology}, 15:201--293, 1904.

\bibitem{android_env}
D.~Toyama, P.~Hamel, A.~Gergely, G.~Comanici, A.~Glaese, Z.~Ahmed, T.~Jackson,
  S.~Mourad, and D.~Precup.
\newblock Androidenv: A reinforcement learning platform for android.
\newblock {\em arXiv preprint arXiv:2105.13231}, 2021.

\bibitem{wang:chi2023}
B.~Wang, G.~Li, and Y.~Li.
\newblock Enabling conversational interaction with mobile ui using large
  language models.
\newblock In {\em Proc. of the 2023 CHI Conference on Human Factors in
  Computing Systems}, CHI '23. Association for Computing Machinery, 2023.

\bibitem{wang2022self}
X.~Wang, J.~Wei, D.~Schuurmans, Q.~Le, E.~Chi, S.~Narang, A.~Chowdhery, and
  D.~Zhou.
\newblock Self-consistency improves chain of thought reasoning in language
  models.
\newblock {\em arXiv preprint arXiv:2203.11171}, 2022.

\bibitem{wei2021finetuned}
J.~Wei, M.~Bosma, V.~Zhao, K.~Guu, A.~W. Yu, B.~Lester, N.~Du, A.~M. Dai, and
  Q.~V. Le.
\newblock Finetuned language models are zero-shot learners.
\newblock In {\em International Conference on Learning Representations}, 2021.

\bibitem{wei2023chainofthought}
J.~Wei, X.~Wang, D.~Schuurmans, M.~Bosma, F.~Xia, E.~Chi, Q.~V. Le, D.~Zhou,
  et~al.
\newblock Chain-of-thought prompting elicits reasoning in large language
  models.
\newblock {\em Advances in neural information processing systems},
  35:24824--24837, 2022.

\bibitem{wu2023politics}
P.~Y. Wu, J.~A. Tucker, J.~Nagler, and S.~Messing.
\newblock Large language models can be used to estimate the ideologies of
  politicians in a zero-shot learning setting.
\newblock {\em arXiv preprint arXiv:2303.12057}, 2023.

\bibitem{yan2023gpt4v}
A.~Yan, Z.~Yang, W.~Zhu, K.~Lin, L.~Li, J.~Wang, J.~Yang, Y.~Zhong, J.~McAuley,
  J.~Gao, et~al.
\newblock Gpt-4v in wonderland: Large multimodal models for zero-shot
  smartphone gui navigation.
\newblock {\em arXiv preprint arXiv:2311.07562}, 2023.

\bibitem{yao2023}
S.~Yao, D.~Yu, J.~Zhao, I.~Shafran, T.~L. Griffiths, Y.~Cao, and K.~Narasimhan.
\newblock Tree of thoughts: Deliberate problem solving with large language
  models.
\newblock {\em arXiv preprint arXiv:2305.10601}, 2023.

\bibitem{yao2022}
S.~Yao, J.~Zhao, D.~Yu, N.~Du, I.~Shafran, K.~R. Narasimhan, and Y.~Cao.
\newblock {ReAct}: Synergizing reasoning and acting in language models.
\newblock In {\em The 11th International Conference on Learning Representations
  (ICLR)}, 2023.

\bibitem{screen-understanding-apple-chi21}
X.~Zhang, L.~de~Greef, A.~Swearngin, S.~White, K.~Murray, L.~Yu, Q.~Shan,
  J.~Nichols, J.~Wu, C.~Fleizach, A.~Everitt, and J.~P. Bigham.
\newblock {Screen Recognition: Creating Accessibility Metadata for Mobile
  Applications from Pixels}.
\newblock In {\em Proc. of the 2021 CHI Conference on Human Factors in
  Computing Systems}, CHI '21, 2021.

\bibitem{zhao2021calibrate}
Z.~Zhao, E.~Wallace, S.~Feng, D.~Klein, and S.~Singh.
\newblock Calibrate before use: Improving few-shot performance of language
  models.
\newblock In {\em International Conference on Machine Learning}, pages
  12697--12706. PMLR, 2021.

\bibitem{zheng2023seeact}
B.~Zheng, B.~Gou, J.~Kil, H.~Sun, and Y.~Su.
\newblock Gpt-4v(ision) is a generalist web agent, if grounded.
\newblock {\em arXiv preprint arXiv:2401.01614}, 2024.

\bibitem{zhou2022least}
D.~Zhou, N.~Sch{\"a}rli, L.~Hou, J.~Wei, N.~Scales, X.~Wang, D.~Schuurmans,
  C.~Cui, O.~Bousquet, Q.~Le, et~al.
\newblock Least-to-most prompting enables complex reasoning in large language
  models.
\newblock {\em arXiv preprint arXiv:2205.10625}, 2022.

\end{thebibliography}

\appendix

\label{sec:appendix}

\section{Forming screen representations}
\label{app:forming-screen-repr}

In this work, a screen representation consists of a textual description of the elements composing the UI. Such screen representation can be derived automatically from developer-provided UI trees or produced using object detection models fine-tuned on UI screens~\cite{screen-understanding-apple-chi21, uied-icse20}. Both approaches have downsides. UI trees (e.g., web DOM or Android accessibility tree) can be noisy, corrupted with missing element descriptions or misaligned structural information~\cite{li2023spotlight}. UI trees are generally very large, especially web DOM trees, and therefore need to be summarized or truncated to fit into an LLM's input. Object detection models can also fail in detecting key on-screen elements. Large vision-language models can also be employed to detect elements in a screen but they tend to hallucinate elements and texts~\cite{yan2023gpt4v}.

In our experiments we use Android Accessibility trees. These contain accessibility-related metadata such as textual descriptions of icons. While accessibility labels provide rich textual information for UI elements they are generally not provided for all elements~\cite{chen20:gui-accessibility}.

Screen representations were formed at each step through a set of heuristics that produced concise lists of natural language descriptions of on-screen elements. We observed that these lists produce similar or better results while requiring much fewer tokens than representations that directly converted UI trees into JSON representations.  We formed these descriptions as follows:

\begin{enumerate}
    \item{Accessibility trees were cleaned by removing any elements and corresponding children which were marked as not visible or had bounding boxes that were completely off the screen. Note that the visibility attribute of elements were not always set correctly and elements that were behind other elements, which can be difficult to detect from the accessibility tree, would not be detected by this method.}
    \item{Accessibility trees were further simplified by removing chains of container elements that had no text, content description or hint text attributes set.}
    \item{A hierarchical list of UI elements was then produced (see examples in \S\ref{app:ui-screens}), with indentation used to indicate the depth of elements in the tree. Natural language descriptions of each element were formed with heuristics.  The heuristics attempted to identify the type of element (e.g., text box, radio button, check box, switch, etc...) based on class names and then formed natural language descriptions incorporating important attributes (e.g., "a check box with the text "include fries" that is not checked") based on the type.  If the type of element was not covered by a specific heuristic, a generic description with the class and package name and important attributes (e.g., text, selection status, etc...) was generated.}
\end{enumerate}

\section{Noisy and partial screen representations}
\label{app:ui-screens}

\begin{figure}
\centering
\includegraphics[width=0.95\textwidth,trim=18 15 115 10,clip]{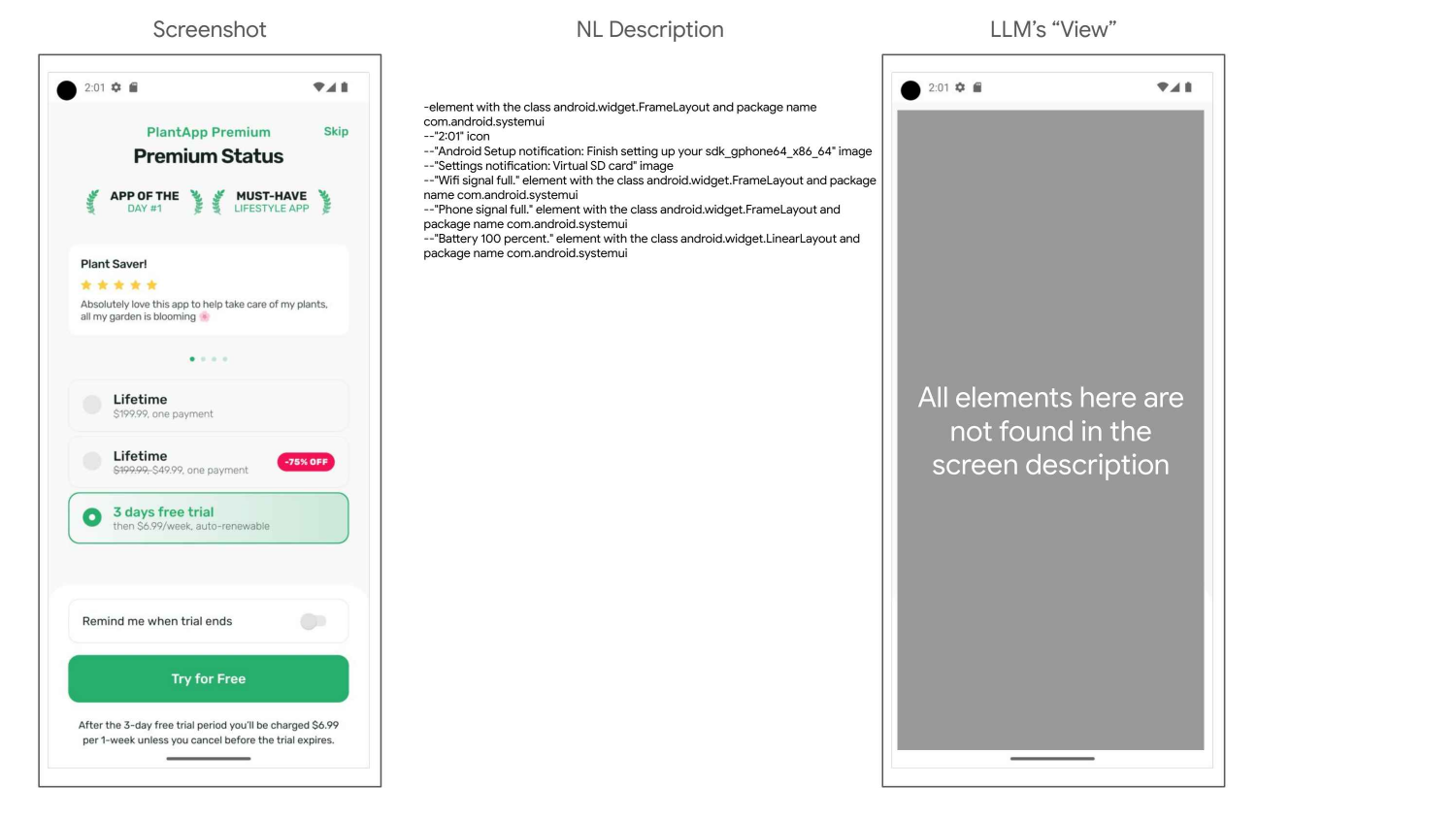}
\caption{Due to the real device environment UI agents run on, we observe cases where the Android accessibility tree becomes out of sync with the true state of the device. In the example above, the current screen (on the left) shows an advertisement for PlantApp Premium. However, because there were delays in the accessibility tree being updated, what is passed to the UI agent is a screen representation missing all the relevant on-screen elements. This results in the agent's ``view'' of the screen looking like the screen on the right, where only the clock and some notification elements are shown.}
\label{fig:sr_example_2}
\end{figure}
\begin{figure}
\centering
\includegraphics[width=0.95\textwidth,trim=5 12 25 10,clip]{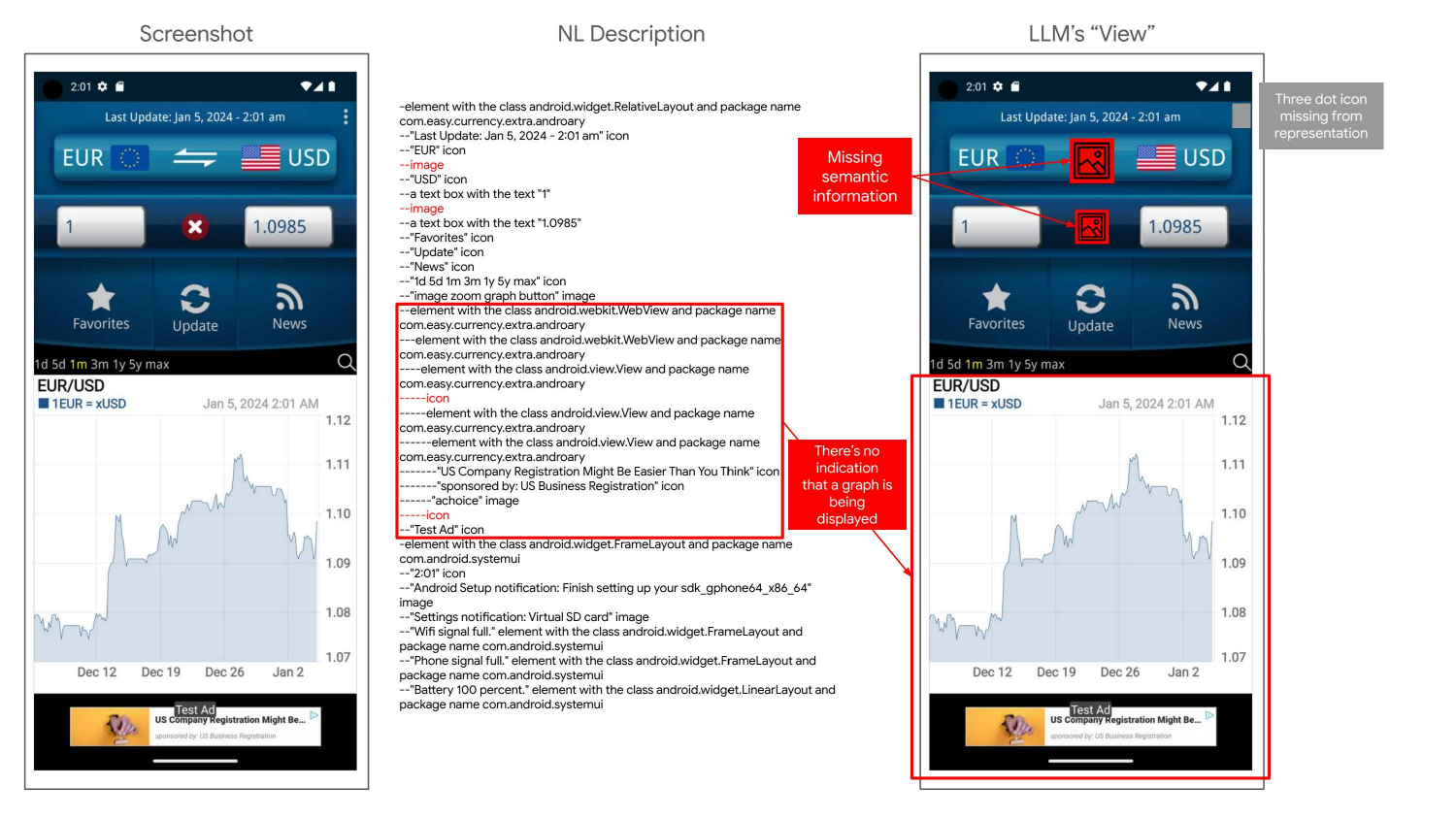}
\caption{As in the previous Figure~\ref{fig:sr_example_2}, this example also contains elements missing textual descriptions that would provide semantic meaning (highlighted in red) and a three-dots icon (in top-right) that is not included in the accessibility tree. To the agent, these elements would look as represented in the screen on the right: red elements as generic image icons and gray elements as non-existent. The graph shown in the screen (which represents half of the page content) appears in the View Hierarchy as a WebView element (blue box), which gives the UI agent no indication that it is a graph and what it is about.}
\label{fig:sr_example_3}
\end{figure}
\begin{figure}
\centering
\includegraphics[width=0.95\textwidth,trim=5 12 25 10,clip]{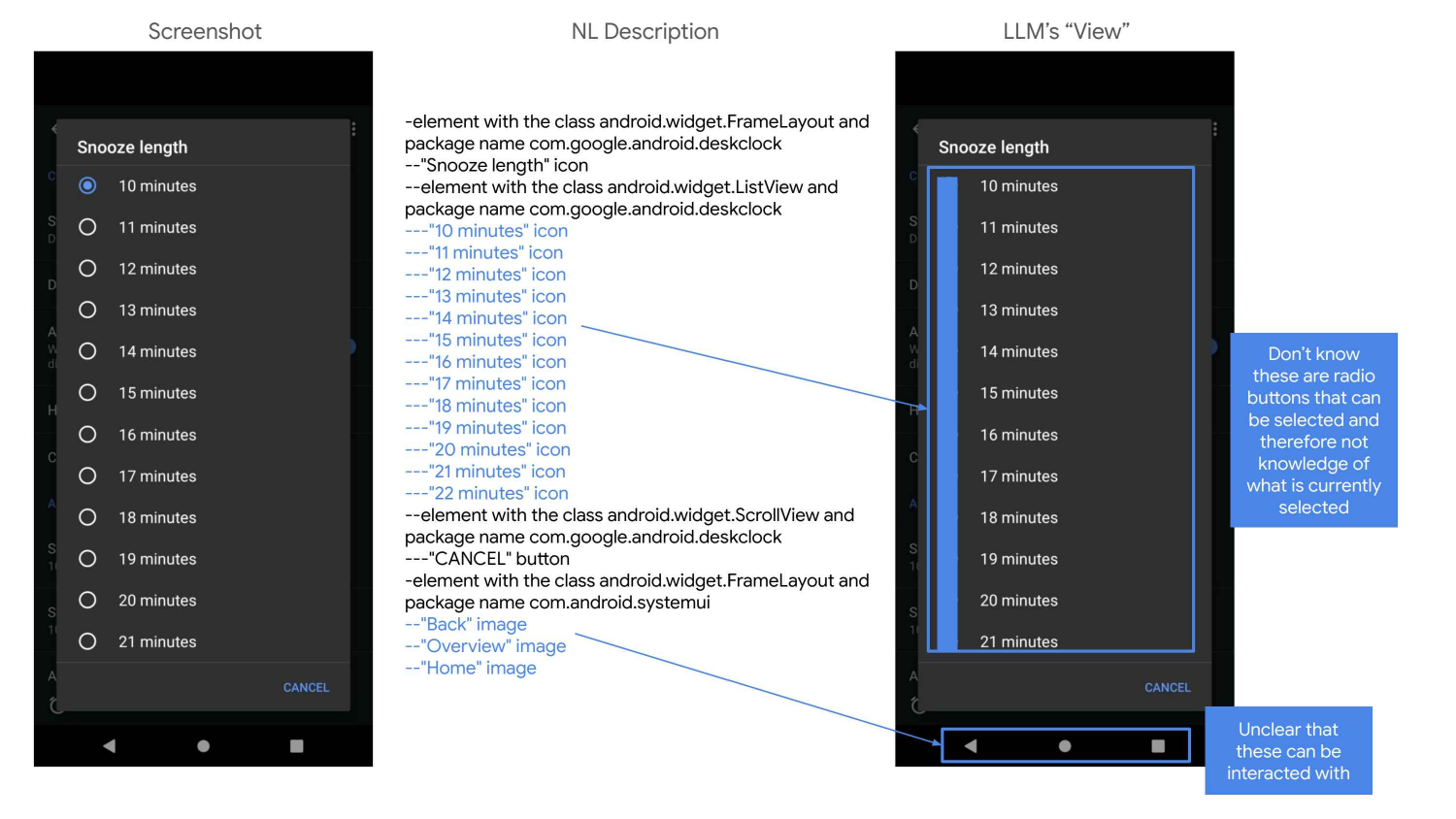}
\caption{The screen on the left shows a dialog box with various options for the snooze length, each option being a selectable radio button. However, if we look at the NL description derived for the screen, these radio buttons are described as icons, giving the model no information that they are selectable elements and therefore no idea about what is currently selected. We also see a similar issue to what was discussed in Figure~\ref{fig:sr_example_2} where the navigation buttons at the bottom of the screen are described in the screen representation as images, so the model may not know they can be interacted with.}
\label{fig:sr_example_5}
\end{figure}
\begin{figure}
\centering
\includegraphics[width=0.95\textwidth,trim=5 10 57 10,clip]{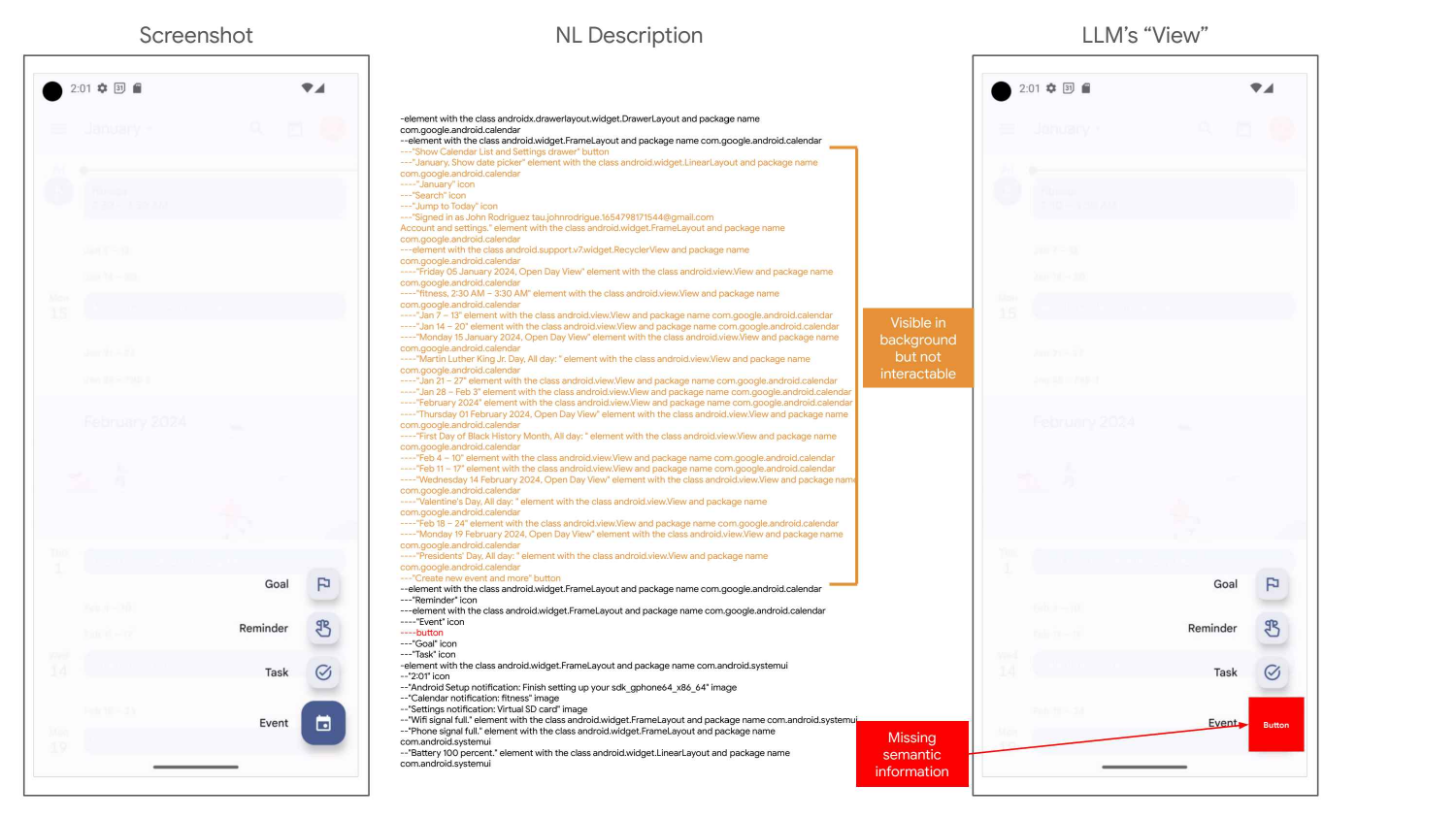}
\caption{This is an example of NL description containing elements that cannot be interacted with. While the elements highlighted in orange can be seen in the background of the screen, they are not clickable. Having these orange-colored elements in the description provides an inaccurate representation to the agent, as it thinks these are elements it can select for next action prediction. This example also contains a button with no text describing its semantic meaning, so the model "sees" the screen on the right, with a generic button, but no direct information about the button's function.}
\label{fig:sr_example_6}
\end{figure}
\begin{figure}
\centering
\includegraphics[width=0.91\textwidth]{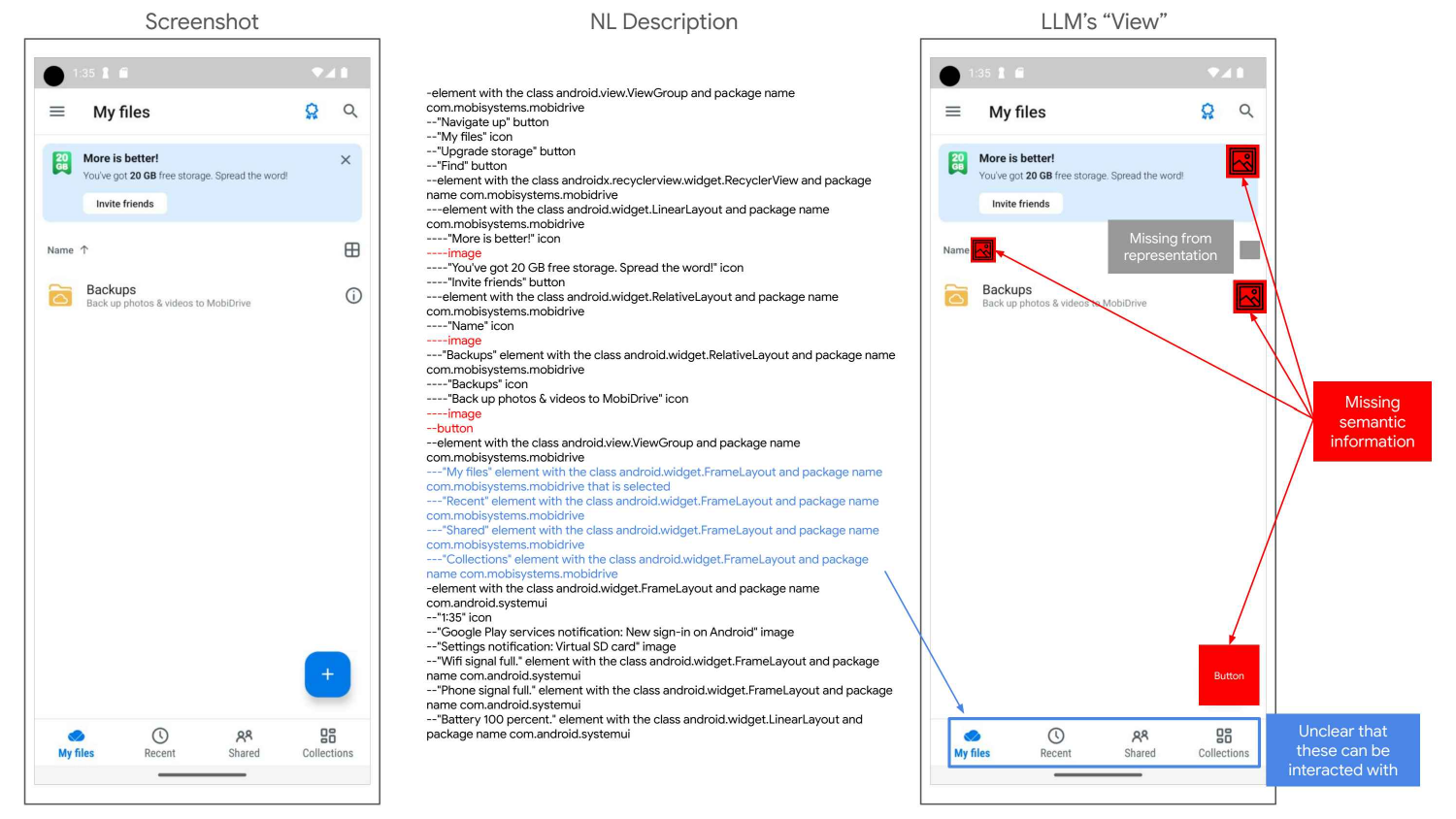}
\caption{There are instances where even if a UI element is present in the accessibility tree, it is missing critical information such as text that would provide semantic meaning for the element or an element type that would tell the agent whether the element is clickable or not. The elements in the NL description that are colored in red do not have any associated text, so it is difficult to know what purpose they have. The agent would view these elements as shown in the right screen, where the elements are known to be an image or button, but what that image shows or that button does is unknown. The elements in blue have text providing semantic meaning, but their types are misleading. The tab elements boxed in blue are labeled as FrameLayout elements, which give no indication that they can be clicked on. Additionally, in this example, there is a grid icon at the center-left of the screen that is not in the tree, so from the agent's perspective it does not exist (shown in the view on the right with a gray box).
}
\label{fig:sr_example_1}
\end{figure}

Figures \ref{fig:sr_example_2}, \ref{fig:sr_example_3}, 
\ref{fig:sr_example_5}, \ref{fig:sr_example_6} and  \ref{fig:sr_example_1} show some example screens as observed by a human (left), the textual representation of that screen as presented to a UI agent (middle) and, for clarity, a visual depiction of the screen high-lighting areas uncertain or missing in the textual representation (right).

\section{The A-50 benchmark}
\label{app:a50}

We collected a third benchmark, ``Android-50,'' consisting of a small set of natural language task instructions. Annotators proficient in English and the use of Android devices were asked to write natural language descriptions of tasks they would accomplish in different task categories, such as creating events, performing unit conversions, and shopping online. We asked annotators to avoid specifying any private information and they received fair compensation. They were also informed about the intended usage of the data. The data collected went through a legal and privacy review.  A sample of tasks written by the annotators can be found in \S\ref{sec:datasets}.

\section{LLM-based grounder}
\label{app:grounder}

\begin{figure}[h]
\centering
\includegraphics[width=\columnwidth]{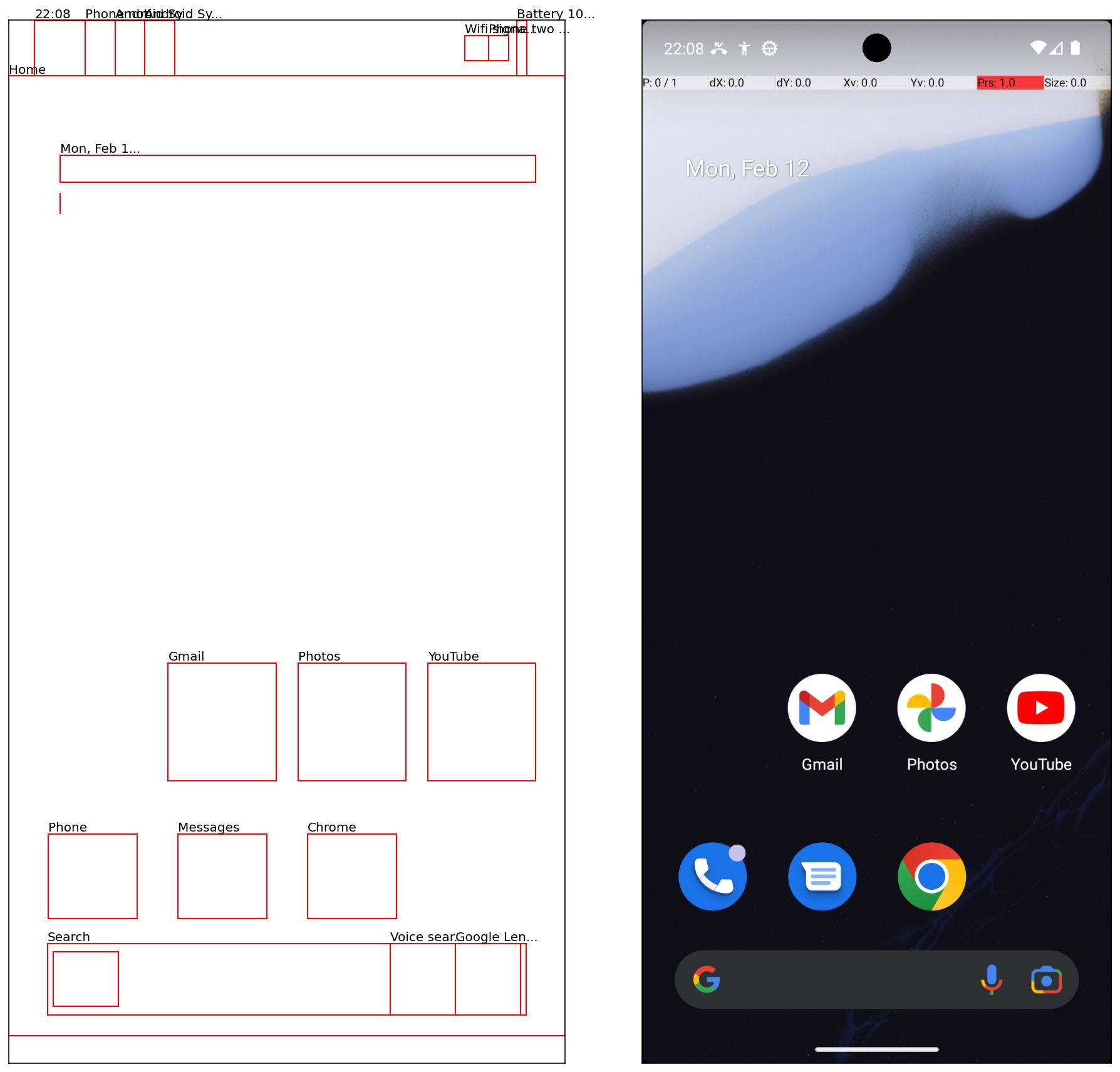}
\vspace{-1ex}
\caption{Visualization of the extracted accessibility nodes for the home screen.}
\label{fig:acc-tree}
\end{figure}

The LLM-based grounder consumes a processed version of the Android accessibility tree, represented as a JSON object, and outputs an action, also encoded in JSON. Specifically, we extract the leaf nodes, along with their text and bounding box properties, the center points and dimensions. For example, below is the representation of the first several elements from the home screen (see also Fig.~\ref{fig:acc-tree}).

\begin{small}
\begin{verbatim}
{"UI elements": [{"text": "Home", "center": [540, 1232], "size": [1080, 2209]}, 
{"text": "Phone", "center": [162, 1970], "size": [173, 195]},
{"text": "Messages", "center": [414, 1970], "size": [173, 195]}, 
{"text": "Chrome", "center": [666, 1970], "size": [172, 195]}, ...]}
\end{verbatim}
\end{small}

Below is the prompt for the grounder, which will output an action. The action is then executed using the Android Device Bridge (adb) (e.g., \texttt{adb shell input tap x y}). The candidate JSON actions are displayed in the prompt.

\begin{small}
\begin{verbatim}
Given a mockup of a mobile interface screen, a
history of past actions, and a desired goal, your
task is to infer the most immediate and logical
action that a user should take to make progress
towards the specified goal. Consider the provided
interface elements, their types, and the text they
contain, as these provide clues about the current
state of the mobile application and potential
actions a user could take. Consider exploring the
screen by scrolling in different directions to
reveal additional content, or using navigation
actions like 'Back' if the current screen does not
contain the elements or options needed to progress
towards the goal.

Actions you can take:

Click on an element on the screen:
'{"action_type": "click", "x": <x_coordinate>,
"y": <y_coordinate>}'.
Type text into a text field '{"action_type":
"input_text", "text": <text_input>, "x":
<x_coordinate>, "y": <y_coordinate>}'.
Press Enter key: '{"action_type":
"keyboard_enter"}'
Navigate to the home screen: '{"action_type":
"navigate_home"}'.
Navigate back: '{"action_type": "navigate_back"}'.
Scroll in a specific direction: '{"action_type":
"scroll", "direction": <up, down, left, or
right>}'.
Open an app: '{"action_type": "open_app",
"app_name": <name>}'.
Launch an ADB activity with a nickname (either
'app_drawer' or 'quick_settings'):
'{"action_type": "launch_adb_activity",
"activity_nickname": <activity_nickname>}'.
Wait for the screen to update: '{"action_type":
"wait"}'.
Note: Always consider the context of the current
screen and the user's goal. If the current screen
does not contain the elements or options needed to
progress towards the goal, consider exploring the
screen by scrolling or using navigation actions
like 'Back'.

Screen:
[Array of UI elements. It includes the text that
describes them, their <x,y> center, and their
width and height. The dimensions of the screen are
1080 x 2400 pixels.]
<SCREEN_REPRESENTATION>

Remember, this mockup is an abstraction of the
mobile interface screen and may not reflect the
exact visual layout. It's intended to highlight
the significant components that could influence
the user's actions.

User Goal: <GOAL>
Choose from the following action types:
```json
[{"action_type": "click", "x": <x_coordinate>,
"y": <y_coordinate>}, {"action_type":
"input_text", "text": <text_input>, "x":
<x_coordinate>, "y": <y_coordinate>},
{"action_type": "keyboard_enter"}, {"action_type":
"navigate_home"}, {"action_type":
"navigate_back"}, {"action_type": "scroll",
"direction": <up, down, left, or right>},
{"action_type": "open_app", "app_name": <name>},
{"action_type": "launch_adb_activity",
"activity_nickname": "<activity_nickname>"}]
```
Answer:
```json

\end{verbatim}
\end{small}

\section{LLM prompts for estimating latent state}
\label{app:latent_state_prompts}

\subsection{Prompt for inferring previous actions}

Below is the prompt that was used for inferring the last action that was performed.  The values of \emph{last\_commanded\_action}, \emph{screen\_1\_description} and \emph{screen\_2\_description} were the agents last action command, and the textual representations of the previous and current screen, respectively.

\begin{small}
\begin{verbatim}
I will show you two screens and tell you
a possible action that I performed to
transition between those two. I may be
lying. Your job is to decide if I am telling
you the real action or not. Only conclude
that I am lying if the evidence cleary
contradicts what I said I did. No matter
what, please tell me the real action you
think I performed. Keep in mind that no
action may have been performed. Please
state the action in the past tense.
Possible action: {last_action_commanded}
Here is a description of the first screen:
{previous_screen_nl_description}
Here is a description of the second screen:
{screen_nl_description}
Real action:
\end{verbatim}
\end{small}

\subsection{Prompt for inferring screen summaries}

Below is the prompt that was used for inferring a high-level summary of the current screen.  Here the value of \emph{screen\_description} was the textual representation (see \S\ref{app:forming-screen-repr}) of the current screen and \emph{last\_inferred\_action} was the last inferred action by the LLM (see prompt above).

\begin{small}
\begin{verbatim}
Here is a description of a screen on an
Android phone:
{screen_description}

Additionally, here is the last action
you took: {last_inferred_action}
What app is this from and what type of screen
is this (e.g., a welcome screen,
advertisement, app screen, etc)? Please
do not describe any past actions.
Only say what app this is and
describe the screen. If media like
a video is showing or playing, please
say what media is showing or playing.
If switches are on the screen, please
say if the the switches are on or off.
\end{verbatim}
\end{small}

\subsection{Prompt for inferring progression}

Below is the prompt that was used for inferring progression.  The value of \emph{inferred\_action\_history\_formatted} was a numbered list of all previous actions (and the message ``Nothing. You are just starting.'' if there were no previous actions). The value of \emph{screen\_summary} was the inferred screen summary above and the \emph{screen\_description} was the textual representation (see \S\ref{app:forming-screen-repr}) of the current screen.

\begin{small}
\begin{verbatim}
I will describe in detail your past actions
towards achieving an unknown goal.  Please
summarize what you have done. Your past
actions were:
{inferred_action_history_formatted}
In addition, here is a summary of the current
screen:
{screen_summary}
Finally, here is a detailed list of elements
on the current screen:
{screen_description}
Without guessing the goal, please summarize
your what you have done and make sure to
mention any specific values that were
entered or selections that were made.
You have
\end{verbatim}
\end{small}

\subsection{Prompt for inferring previous mistakes}

Below is the prompt for inferring if any mistakes have been made.  The value of \emph{cleaned\_goal} is a normalized version of the spoken goal command, where an LLM is asked to put potentially grammatically incorrect goal commands in standard English. We treat this goal normalization as part of the agent, and it was kept the same across all experiments.  The value of \emph{screen\_description} was the textual representation (see \S\ref{app:forming-screen-repr}) of the current screen.

\begin{small}
\begin{verbatim}
I asked you to use an Android phone
to {cleaned_goal}Here is a summary of
your progress so far: {progress_summary}
Here is a description of the current screen:
{screen_description}

If no mistakes have been made, please
simply say "No mistakes have been made."
Otherwise, please tell me if you need
to correct anything to achieve the goal.
\end{verbatim}
\end{small}

\subsection{Prompt for inferring task completion}
\label{sec:completion_prompt}

Below is the prompt for inferring if a task is complete.  The \emph{cleaned\_goal} is a normalized version of the goal command (see above).  The value of \emph{inferred\_action\_history\_formatted} was a numbered list of all previous actions (and the message ``Nothing. You are just starting.'' if there were no previous actions). The \emph{screen\_summary} was provided by the inferred screen summary (see above) and the \emph{possible\_action\_command} was the action that the reasoning method for selecting the next step (see \S\ref{app:action_selection}) predicted. Because we only provided inferences of the progression and mistake aspects of latent state to the LLM when predicting next actions, it was possible to predict next actions before inferring if the task was complete.  We found that providing the LLM with a contemplated next action helped improve its accuracy at determining if a task was done. If it was inferred that the task was complete, the agent stopped instead of performing the next predicted action.

\begin{small}
\begin{verbatim}
I asked you to use an Android phone
to {cleaned_goal}
Here is a list of your past actions:
{inferred_action_history_formatted}
Here is a summary of the screen you
currently see: {screen_summary}
Here is a next step you could perform:
{possible_action_command} However, you
should only do only do the minimum
number of steps needed to achieve
your goal and you should not do anything
if the goal is already achieved.
Yes or No, have you completed all
required steps?
\end{verbatim}
\end{small}

\section{Additional details of agent architecture and methods for selecting next actions}
\label{app:action_selection}

The main details of agent architecture we used in this paper are provided in \S\ref{sec:methods}.  We now provide a few additional details, space did not permit to be included in the main text.

\paragraph{Goal command normalization} For all experiments, we performed a form of normalization of the goal command by passing it through an LLM with the following prompt before passing it to any further agent LLM calls:

\begin{small}
\begin{verbatim}
Here are some examples of how requests can
be rephrased into proper imperative sentences:

Request: Read the reviews of Audi Q7 in
CarTrade app
Read the reviews for the Audi Q7 in the
CarTrade app

Request: In Meditopia app, Play "Thunder
and Rain" audio in Sleep mode.
Rephrased: In the Meditopia app, play the
"Thunder and Rain" audio in sleep mode.'

Request: Open the EaseMyTrip app and
search for the Le Roi Express
Hotel in Delhi from August 1–23 to August 5–23
Rephrased: Open the EaseMyTrip app and
search for the Le Roi Express
Hotel in Delhi for the dates
August 1, 2023 to August 5, 2023.

Here is a new request: <original_request>'
Please rephrase it.
Sure, here is how that request can
be rephrased:
\end{verbatim}
\end{small}

\paragraph{Parsing action commands}

While output of the LLM call to predict next actions was of a form that could be immediately sent to the grounder, the output with the Cot-SC and \react~based methods needed to be stripped of intermediate thoughts. For the CoT-SC agents, this was done by returning everything after the delimiter 'Answer:' if the delimiter was found and the last sentence if not.
For the \react~based agents, this was done by application of a simple regular expression looking for a delimited thought and action and returning ``unknown'' if a matching pattern could not be found.

\paragraph{Determining when to stop}

Agents equipped with latent state estimation benefited from explicit estimates of task completion (\S\ref{sec:latent_state_for_ui_automation}).  These agents stopped performing a task whenever the response to the prompt for inferring completion (\S\ref{sec:completion_prompt}) began with the string "Yes."

Agents without the benefit of latent state estimation predicted stopping as another type of action directly in their action selection step.  Similar to agents with latent state, we determine stopping for these by detecting the string "done" in the predicted action.

\subsection{Zero-shot -}

Below is the prompt that was used for predicting next actions when estimates of latent state were not provided for the zero-shot- agent. The value of \emph{cleaned\_goal} was the normalized version of the goal command (see discussion of general agent architecture above). The \emph{formatted\_history\_of\_commanded\_actions} was a numbered list of previous action commands (and the message ``1) None.'' if there were no previous actions). The \emph{screen\_description} was the textual representation (see \S\ref{app:forming-screen-repr}) of the current screen.

\begin{small}
\begin{verbatim}
I asked you to use an Android phone
to {goal_clean}
Here are the actions you have taken
so far:
{formatted_history_of_commanded_actions}
Here is a detailed description
of the current screen:
{screen_description}
What is the next action that you
should take? Remember you can not
interact with something if it is
not visible on the screen. However,
you can open an app even if you do
not see it. To open an app just say
to open it. What is the next action
that you should take? Please state
it succinctly and only state one
step at a time. If you have accomplished
the goal, simply state "You should
be done." You should
\end{verbatim}
\end{small}

\subsection{Zero-shot +}

Below is the prompt that was used for predicting next actions when estimates of latent state were provided for the zero-shot+ agent. The value of \emph{cleaned\_goal} was the normalized version of the goal command (see discussion of general agent architecture above).  The values of \emph{mistake\_assessment} and \emph{progression} were the latent state estimates of those quantities.  The \emph{screen\_description} was the textual representation (see \S\ref{app:forming-screen-repr}) of the current screen. To ensure as close to a head-to-head comparison as possible the prompt has been minimally modified relative to the prompt for the zero-shot- agent.  Specifically, only the addition of injecting estimates of latent state into the prompt and removing the instruction to predict stopping was change. As explained above, stopping for agents with latent state was based directly on the estimation of task completion.

\begin{small}
\begin{verbatim}
I asked you to use an Android phone
to {cleaned_goal}
Here is a summary of your progress
so far: You have {progression}
Here are mistakes that need to be
corrected: {mistake_assessment}
Here is a detailed description
of the current screen:
{screen_description}
What is the next action that you
should take? Remember you can not
interact with something if it is
not visible on the screen. However,
you can open an app even if you do
not see it. To open an app just say
to open it. What is the next action
that you should take? Please state
it succinctly and only state one
step at a time. You should
\end{verbatim}
\end{small}

\subsection{CoT-SC -}

Below is the prompt that was used for predicting next actions when latent state estimates were not provided for the CoT-SC- agent.  Below is the prompt that was used for predicting next actions when latent state estimates were not provided for the CoT-SC- agent.  For brevity below, the full textual screen descriptions that were provided in the prompt have been replaced in the presentation below by the place holders "[example\_n\_screen\_description]."   The values of \emph{cleaned\_goal}, \emph{formatted\_history\_of\_commanded\_actions} and \emph{screen\_description} were the same as the prompt for the zero-shot- agent (see above).

\begin{small}
\begin{verbatim}
I asked you to use an Android phone 
to Use color inversion.
Here are the actions you have taken 
so far:
1) open settings
Here is a detailed description 
of the current screen:
[example_1_screen_description]
What is the next action that you 
should take? Remember you can not 
interact with something if it is 
not visible on the screen. However, 
you can open an app even if you 
do not see it. To open an app just 
say to open it. What is the next 
action that you should take? Please 
state it succinctly and only state 
one step at a time. If you have 
accomplished the goal, simply state 
"You should be done."
Let's think step by step. The color 
inversion setting is probably under 
an accessibility menu.  I do not 
see accessibility on the current 
screen, but I can probably use 
the search box to search for it.
 Answer: click on the search box

I asked you to use an Android phone 
to Set an alarm for 3:30 PM with 
the alarm sound Argon.
Here are the actions you have taken 
so far:
1) open Clock
Here is a detailed description 
of the current screen:
[example_2_screen_description]
What is the next action that you 
should take? Remember you can not 
interact with something if it is 
not visible on the screen. However, 
you can open an app even if you 
do not see it. To open an app just 
say to open it. What is the next 
action that you should take? Please 
state it succinctly and only state 
one step at a time. If you have 
accomplished the goal, simply state 
"You should be done."
Let's think step by step. I have 
opened the clock app but I am on 
the wrong tab. I can correct that 
by clicking on the alarm tab. Answer: 
tap the alarm tab.

I asked you to use an Android phone 
to In Simple Calendar Pro, create 
a task titled "Exercise" with the 
description "Run for 30 minutes" 
for today.
Here are the actions you have taken 
so far:
1) open Simple Calendar Pro
2) click the New Event button
3) click the task icon
4) type Exercise in the Title text 
box
Here is a detailed description 
of the current screen:
[example_3_screen_description]
What is the next action that you 
should take? Remember you can not 
interact with something if it is 
not visible on the screen. However, 
you can open an app even if you 
do not see it. To open an app just 
say to open it. What is the next 
action that you should take? Please 
state it succinctly and only state 
one step at a time. If you have 
accomplished the goal, simply state 
"You should be done."
Let's think step by step. I have 
opened the Simple Calendar Pro 
app and have started to create 
a new task. I added the title without 
making any mistakes.  Now I need 
to add a description.  I see a 
description text box, which I can 
use for this. Answer: type 'Run 
for 30 minutes' in the Description 
text box.

I asked you to use an Android phone 
to {cleaned_goal}
Here are the actions you have taken 
so far:
{formatted_commanded_action_history}
Here is a detailed description 
of the current screen:
{screen_description}
What is the next action that you 
should take? Remember you can not 
interact with something if it is 
not visible on the screen. However, 
you can open an app even if you 
do not see it. To open an app just 
say to open it. What is the next 
action that you should take? Please 
state it succinctly and only state 
one step at a time. If you have 
accomplished the goal, simply state 
"You should be done."
Let's think step by step.
\end{verbatim}
\end{small}

\subsection{CoT-SC +}

Below is the prompt that was used for predicting next actions when latent state estimates were provided for the CoT-SC~+ agent.  For brevity below, the full textual screen descriptions that were provided in the prompt have been replaced in the presentation below by the place holders "[example\_n\_screen\_description]."   The values of \emph{cleaned\_goal}, \emph{progress\_summary}, \emph{mistake\_assessment} and \emph{screen\_description} are the same as they were for the zero-shot+ agent (see above).

\begin{small}
\begin{verbatim}
I asked you to use an Android phone 
to Use color inversion.
Here is a summary of your progress 
so far: You have opened the settings 
app.
Here are mistakes that need to 
be corrected: No mistakes have 
been made.
Here is a detailed description 
of the current screen:
[example_1_screen_description]
What is the next action that you 
should take? Remember you can not 
interact with something if it is 
not visible on the screen. However, 
you can open an app even if you 
do not see it. To open an app just 
say to open it. What is the next 
action that you should take? Please 
state it succinctly and only state 
one step at a time.
Let's think step by step. The color 
inversion setting is probably under 
an accessibility menu.  I do not 
see accessibility on the current 
screen, but I can probably use 
the search box to search for it.
 Answer: click on the search box

I asked you to use an Android phone 
to Set an alarm for 3:30 PM with 
the alarm sound Argon.
Here is a summary of your progress 
so far: You have opened the Clock 
app and selected the Bedtime tab.
Here are mistakes that need to 
be corrected: You are on the Bedtime 
tab. You need to be on the Alarm 
tab. Click the Alarm tab.
Here is a detailed description 
of the current screen:
[example_2_screen_description]
What is the next action that you 
should take? Remember you can not 
interact with something if it is 
not visible on the screen. However, 
you can open an app even if you 
do not see it. To open an app just 
say to open it. What is the next 
action that you should take? Please 
state it succinctly and only state 
one step at a time.
Let's think step by step. I have 
opened the clock app but I am on 
the wrong tab. I can correct that 
by clicking on the alarm tab. Answer: 
tap the alarm tab.

I asked you to use an Android phone 
to In Simple Calendar Pro, create 
a task titled "Exercise" with the 
description "Run for 30 minutes" 
for today.
Here is a summary of your progress 
so far: You have opened Simple 
Calendar Pro and clicked the New 
Event button. You then clicked 
the task icon and typed Exercise 
in the Title text box.
Here are mistakes that need to 
be corrected: No mistakes have 
been made.
Here is a detailed description 
of the current screen:
[example_3_screen_description]
What is the next action that you 
should take? Remember you can not 
interact with something if it is 
not visible on the screen. However, 
you can open an app even if you 
do not see it. To open an app just 
say to open it. What is the next 
action that you should take? Please 
state it succinctly and only state 
one step at a time.
Let's think step by step. I have 
opened the Simple Calendar Pro 
app and have started to create 
a new task. I added the title without 
making any mistakes.  Now I need 
to add a description.  I see a 
description text box, which I can 
use for this. Answer: type 'Run 
for 30 minutes' in the Description 
text box.

I asked you to use an Android phone 
to {cleaned_goal}
Here is a summary of your progress 
so far: You have {progress_summary}
Here are mistakes that need to 
be corrected: {mistake_assessment}
Here is a detailed description 
of the current screen:
{screen_description}
What is the next action that you 
should take? Remember you can not 
interact with something if it is 
not visible on the screen. However, 
you can open an app even if you 
do not see it. To open an app just 
say to open it. What is the next 
action that you should take? Please 
state it succinctly and only state 
one step at a time.
Let's think step by step.
\end{verbatim}
\end{small}

\subsection{\react~-}
\label{sec:react_minus_prompt}
Below is the prompt that was used for predicting next actions when estimates of latent state were not provided for the \react~- agent.  The values of \emph{cleaned\_goal} and \emph{screen\_description} are the same as they were for the zero-shot+ agent (see above). The value of \emph{observation\_thought\_action\_history} is the full history of thoughts and actions retaining only the last two previous observations (of the form ``Observation 1: ... Thought 1: ... Action 1: ... Observation 2: ...'' with each observation, thought and action starting on separate lines).

\begin{small}
\begin{verbatim}
Here are some examples of thoughts 
and actions:
Thought: I just opened settings.
 The color inversion setting is 
probably under an accessibility 
menu.  I do not see accessibility 
on the current screen, but I can 
probably use the search box to 
search for it
Action: click on the search box
Thought: I have opened the clock 
app but I am on the wrong tab. 
I can correct that by clicking 
on the alarm tab.
Action: tap the alarm tab.
Thought: I have opened the Simple 
Calendar Pro app and have started 
to create a new task. I added the 
title without making any mistakes.
  Now I need to add a description.
  I see a description text box, 
which I can use for this.
Action: type "Run for 30 minutes" 
in the Description text box.

I asked you to use an Android phone 
to {cleaned_goal}
Here is a history of previous screen 
descriptions, thoughts and actions:
{observation_thought_action_history}
The actions above may have had 
unexpected results, so you should 
pay special attention to the current 
screen below.
Here is a detailed description 
of the current screen:
{screen_description}
Please think about what you should 
do next and then pick the next 
action that you should take.
Your output should be of the form:
Thought: <thought>
Action: <action>
Please state the action succinctly 
and only state one step at a time.
Remember you can not interact with 
something if it is not visible 
on the screen, so you must look 
carefully at the current screen 
to make sure your action can be 
performed on it. However, you can 
open an app even if you do not 
see it. To open an app just say 
to open it. If you have accomplished 
the goal, you should think about 
why you are done and then the action 
should just be "done."
Thought: 
\end{verbatim}
\end{small}

\subsection{\react~+}
\label{sec:react_plus_prompt}
Below is the prompt that was used for predicting next actions when latent state estimates were provided for the\react~+ agent.  The values of \emph{cleaned\_goal}, \emph{progress\_summary}, \emph{mistake\_assessment} and \emph{screen\_description} are the same as they were for the zero-shot+ agent (see above). The value of \emph{observation\_thought\_action\_history} was formed as in the \react~- agent.

\begin{small}
\begin{verbatim}
Here are some examples of thoughts 
and actions:
Thought: I just opened settings.
 The color inversion setting is 
probably under an accessibility 
menu.  I do not see accessibility 
on the current screen, but I can 
probably use the search box to 
search for it
Action: click on the search box
Thought: I have opened the clock 
app but I am on the wrong tab. 
I can correct that by clicking 
on the alarm tab.
Action: tap the alarm tab.
Thought: I have opened the Simple 
Calendar Pro app and have started 
to create a new task. I added the 
title without making any mistakes.
  Now I need to add a description.
  I see a description text box, 
which I can use for this.
Action: type "Run for 30 minutes" 
in the Description text box.

I asked you to use an Android phone 
to {cleaned_goal}
Here is a summary of your progress 
so far: You have {progress_summary}
Here are mistakes that need to 
be corrected: {mistake_assessment}
Here is a history of previous screen 
descriptions, thoughts and actions:
{observation_thought_action_history}
The actions above may have had 
unexpected results, so you should 
pay special attention to the current 
screen below.
Here is a detailed description 
of the current screen:
{screen_description}
Please think about what you should 
do next and then pick the next 
action that you should take.
Your output should be of the form:
Thought: <thought>
Action: <action>
Please state the action succinctly 
and only state one step at a time.
Remember you can not interact with 
something if it is not visible 
on the screen, so you must look 
carefully at the current screen 
to make sure your action can be 
performed on it. However, you can 
open an app even if you do not 
see it. To open an app just say 
to open it.
Thought: 
\end{verbatim}
\end{small}

\section{UI agent benchmarks}
\label{sec:datasets}

We present a subset of 10 tasks from each of the benchmarks that we used for evaluation in Table \ref{tab:benchmark_tasks}.

\begin{table*}[!ht]
\centering
\caption{A subset of 10 task goals from each of our test benchmarks.}
\scalebox{0.8}{
\begin{tabular}{l|l}
    \toprule
    \textbf{Benchmark} & \textbf{Task goal}\\
    \midrule
    \multirow[c]{10}{*}{PixelHelp}
    & Change the clock display to digital \\
    \cline{2-2}
    & Turn notification dots on \\
    \cline{2-2}
    & Turn off data roaming \\
    \cline{2-2}
    & Open a new tab in the chrome app \\
    \cline{2-2}
    & Change alarm snooze length \\
    \cline{2-2}
    & Stop showing notifications on the lock screen \\
    \cline{2-2}
    & Refresh tabs in the chrome app \\
    \cline{2-2}
    & Check data usage \\
    \cline{2-2}
    & Show emergency info \\
    \cline{2-2}
    & Turn on private dns \\
    \midrule
    \multirow[c]{10}{*}{AitW}
    & Install app "Walmart Shopping \& Grocery" \\
    \cline{2-2}
    & Open app "Skype" (install if not already installed) and go to login screen \\
    \cline{2-2}
    & Open app "Google Keep" (install if not already installed) \\
    \cline{2-2}
    & Install app "eBay: The shopping marketplace" \\
    \cline{2-2}
    & Play the latest video from the Wall Street Journal \\
    \cline{2-2}
    & Install app "McDonald's" \\
    \cline{2-2}
    & Clear all items from cart on walmart.com. Add usb-a to usb-b to the cart on walmart.com, then select checkout. \\
    \cline{2-2}
    & Open calendar and show me the second week of next month \\
    \cline{2-2}
    & Add acer predator to the cart on costco.com, then select checkout. \\
    \cline{2-2}
    & Open app "VLC for Android" (install if not already installed) \\
    \midrule
    \multirow[c]{10}{*}{ A-50 }
    & Open the file manager app and create a folder named as My Pictures \\
    \cline{2-2}
    & Open my most recent message in Google Chat. \\
    \cline{2-2}
    & Open the Artier app and search for art by Salvador Dali. \\
    \cline{2-2}
    & Open the PlantApp and create a garden plan for foxtail fern including seeds, water, and light required.  \\
    \cline{2-2}
    & Go to the Fit app and read how much sleep you need from the AASM \\
    \cline{2-2}
    & Go to the Kitchen Stories app and save the recipe for spaghetti  \\
    \cline{2-2}
    & Open the Any.do app and set a reminder for 2:30 PM for the Pilates Classes task. \\
    \cline{2-2}
    & Open the MuniMobile app, go to Security setting change the password  \\
    \cline{2-2}
    & In the Fresh to home app, add the product premium chicken skinless to the cart . \\
    \cline{2-2}
    & Open MobiDrive app and search for the file named as PXL\_20230622\_091318450.jpg \\
    \midrule
\end{tabular}
}
\label{tab:benchmark_tasks}
\end{table*}

\section{Assessing accuracy of latent state estimation}
\label{app:latent_state-questions}

We used a pool of English-speaking annotators, proficient in the use of Android devices to assess the accuracy of the latent state estimates.  The annotators were provided written instructions, one hour of training on how to perform the annotation, and written feedback on their annotations.

\begin{figure*}
\centering
\includegraphics[width=0.95\textwidth,trim=5 5 5 5,clip]{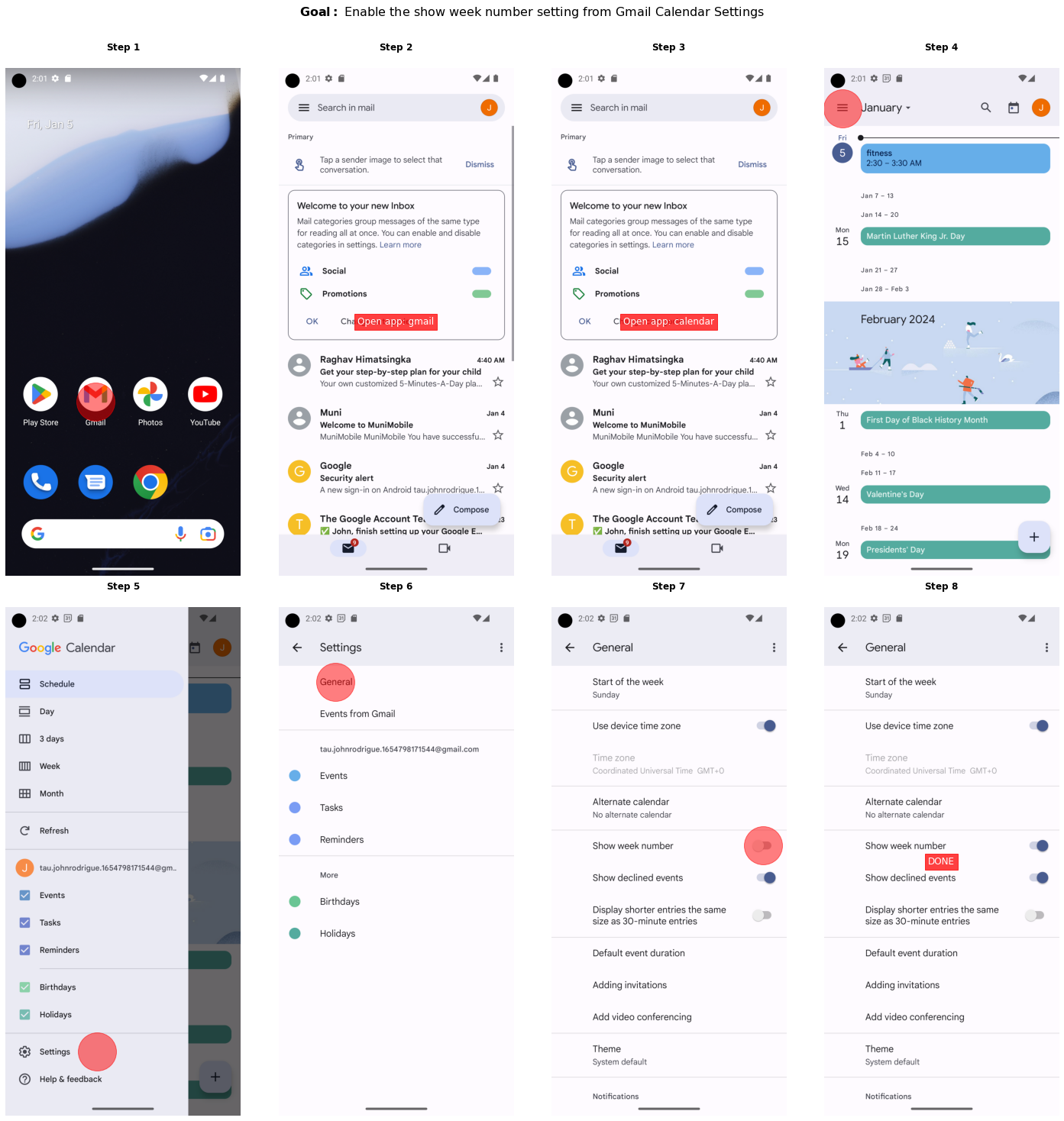}
\caption{An example visualization that we provide to the annotators when evaluating the accuracy of latent state estimates. At the top we provide the task's goal and for each step we provide a screenshot with a visualization of the action plotted on top.}
\label{fig:latent_state_visualization}
\end{figure*}

To perform the annotation, annotators were provided with screenshots of each step in the task an agent performed as well as visualizations of the performed action (see Fig.~\ref{fig:latent_state_visualization}), and asked to complete a form for each task an agent performed. The form had multiple pages, each corresponding to one step in a task, in which we asked the annotators to answer the following questions based on the visualization of the agents performance:

\begin{itemize}
    \item Does the summary of the screen for Step n provide an accurate high-level description?
    \item Does the screen summary for Step n include any incorrect information?
    \item Does the screen summary for Step n include any incorrect information, disregarding incorrect position information?
    \item Was the action that caused the transition between Step n-1 and Step n inferred correctly?
    \item Does the mistake assessment correctly capture the mistakes (if any) that have been made up to Step n and are not yet corrected?
    \item Does the progression description accurately summarize what has been performed up to Step n?
    \item At Step n, did the planner correctly determine if the goal had been completed yet or not?
    \item Was the action between Step n and Step n+1 a mistake? (Note: This question is not asked for the last step in the episode.)
\end{itemize}

\section{Assessing end-to-end performance}
\label{app:rater-questions}

We used a pool of English-speaking annotators, proficient in the use of Android devices to assess the end-to-end accuracy of the agents.  The annotators were provided written instructions and written feedback on their annotations.

To perform the annotation, the annotators were provided with screenshots of each step in the task an agent performed as well as visualizations of the performed action, and asked to complete a form for each task an agent performed. The form was a single page in which we asked the annotators to answer the following questions based on the visualization:

\begin{itemize}
    \item Regardless of whether it stopped at the right time or not, did the agent successfully complete the task?
    \item A series of partial completion questions specific to the task (see Table \ref{tab:rater-questions} for examples).
    \item Did the agent correctly predict that it was done with the task? \\\\
    The possible responses were:
    \begin{itemize}
        \item Yes, the agent stopped at the right time.
        \item No, the agent stopped prematurely.
        \item No, the agent took extra steps before stopping.
        \item No, the agent did not stop.
    \end{itemize}
\end{itemize}

\begin{table}[!ht]
\centering
\caption{Examples of partial completion questions.}
\scalebox{0.92}{
\begin{tabular}{p{0.47\textwidth}|p{0.55\textwidth}}
    \toprule
    \textbf{Task goal} & \textbf{Questions}\\
    \midrule
    \multirow[c]{4}{*}{Stop showing notifications on the lock screen}
    & Was the settings app opened? \\
    \cline{2-2}
    & Was the notifications page of settings reached? \\
    \cline{2-2}
    & At the end of the episode, were notifications set to not be shown on the lock screen? \\
    \midrule
    \multirow[c]{3}{*}{Check data usage}
    & Was the settings app opened? \\
    \cline{2-2}
    & Was the data usage page of settings reached? \\
    \cline{2-2}
    & Was data usage displayed? \\
    \midrule
    \multirow[c]{4}{*}{Play the latest video from the Wall Street Journal}
    & Was the YouTube app opened? \\
    \cline{2-2}
    & Was a search for a Wall Street Journal video performed? \\
    \cline{2-2}
    & Was a Wall Street Journal video played? \\
    \cline{2-2}
    & Was the latest Wall Street Journal video played? \\
    \midrule
    \multirow[c]{8}{0.45\textwidth}{Clear all items from cart on walmart.com. Add usb-a to usb-b to the cart on walmart.com, then select checkout.}
    & Was the Chrome app opened? \\
    \cline{2-2}
    & Was the walmart.com page reached? \\
    \cline{2-2}
    & Was the cart reached? \\
    \cline{2-2}
    & Was the cart cleared if it needed to be (answer yes if it was not cleared but did not need to be)? \\
    \cline{2-2}
    & Was a usb-a to usb-b cord searched for? \\
    \cline{2-2}
    & Was something added to the cart? \\
    \cline{2-2}
    & Was checkout selected? \\
    \midrule
    \multirow[c]{3}{0.45\textwidth}{Go to the Kitchen Stories app and save the recipe for spaghetti }
    & Was the Kitchen Stories app opened? \\
    \cline{2-2}
    & Was the recipe for spaghetti reached? \\
    \cline{2-2}
    & Was the recipe for spaghetti saved? \\
    \midrule
    \multirow[c]{2}{0.45\textwidth}{Open the MuniMobile app, Go to Security setting change the password }
    & Was the MuniMobile app opened? \\
    \cline{2-2}
    & Was the change password page reached? \\
    \bottomrule
\end{tabular}
}
\label{tab:rater-questions}
\end{table}

\section{Explanation of accuracy of na\"ive latent state methods}
\label{app:naive-method}

An agent that always predicts that a task is incomplete will be correct at every step aside from those where the task has been completed. To determine the cases where the task stands complete, we parse our results for the steps where the agent predicted the task was complete and the annotator said that was correct or the agent predicted the task was incomplete and the annotator said that was incorrect. These two scenarios represent 7.0\% of our data, which means that an agent that always predicts that a task is incomplete would have a task completion accuracy of 92.8\%. From these 7\% of cases where the task is truly complete, our agent properly predicts that the task is complete 81.0\% of the time.

An agent that predicts the previous performed action to always match the commanded action for that step would only be wrong in the situation where the last performed action was not what was commanded (e.g., the grounder did not properly perform the action). To find the cases where this happens, we consider the following four possibilities:
\begin{enumerate}
    \item The inferred last action matches the commanded action and the annotator said the inferred last action was correct.
    \item The inferred last action matches the commanded action and the annotator said the inferred last action was incorrect.
    \item The inferred last action does not match the commanded action and the annotator said the inferred last action was correct.
    \item The inferred last action does not match the commanded action and the annotator said the inferred last action was incorrect.
\end{enumerate}
Possibilities 2 and 3 represent the scenarios where the last action is not the commanded action for that step. To determine if the inferred last action matched the commanded action, we use the fuzzywuzzy\footnote{\url{https://pypi.org/project/fuzzywuzzy/}} Python library (under a GNU General Public License v2 ) to calculate a simple ratio between the two strings, considering the two a match if the ratio was above 0.8 and "do not" was not present in the inferred last action (occasionally ``do not \{commanded action\}'' would be predicted). These possibilities represent 14.2\% of our data, so the agent that predicts the previous performed action to always match the commanded action for that step would have an inferred last action accuracy of 85\%. Of the cases where the last action was not the commanded action, our agent properly inferred the last action 61.3\% of the time.

An agent that predicts there is always no mistake in the previous steps would be wrong in the cases where a mistake had occurred previously and not been corrected. We can determine these cases by considering the scenarios where the agent said no mistakes had occurred and our annotators marked it as wrong, or it output something else and our annotators said it was correct. These two scenarios represent 25.94\% of our steps, which means the na\"ive agent would have a success rate of 74.06\%. Of these cases, our agent correctly captured the mistakes in its assessment 26.32\% of the time.

\end{document}